%% file: main.tex
\RequirePackage{xcolor}
\documentclass{ar-1col-S2O}

\input{includes}

\title{Sampling-Based Motion Planning: A Comparative Review}
\author{Andreas Orthey and Constantinos Chamzas and Lydia E. Kavraki}

\begin{document}
\input{src/00_abstract}

\maketitle

\input{src/01_introduction}
\input{src/03_history}
\input{src/04_algorithms}

\input{src/05_extensions}
\input{src/06_competitions}
\input{src/07_evaluations}
\input{src/08_conclusion}
\input{src/09_relatedwork}

\vspace*{-0.4cm}
\section*{ACKNOWLEDGMENTS}
Work on this article by Lydia E. Kavraki is supported in part by NSF RI 2008720. 
\vspace*{-0.4cm}

\bibliographystyle{ar-style3}
{\footnotesize
\bibliography{general}
}

\end{document}

%% file: includes.tex
\usepackage[utf8]{inputenc}
\usepackage{balance}
\usepackage{hyperref}
\usepackage{makecell}
\usepackage{tabularx}
\usepackage{xinttools}
\usepackage[style=base, font=small]{subcaption}
\usepackage{rotating}
\usepackage{xspace}
\usepackage{xcolor}

\usepackage[free-standing-units=true]{siunitx}
\usepackage{wrapfig}
\usepackage{enumitem}

\usepackage{amsmath}
\usepackage{amsfonts}
\usepackage[numbers]{natbib}
\usepackage{soul} 

\usepackage[noend]{algpseudocode}
\usepackage{algorithm}

\newcolumntype{V}{>{\centering\arraybackslash}m{.033\linewidth}}
\newcolumntype{Z}{>{\raggedleft\arraybackslash}m{.01\linewidth}}
\newcolumntype{Y}{>{\centering\arraybackslash}X}
\newcolumntype{+}{!{\vrule width 1.2pt}}

\newcommand{\R}{\mathbb{R}}
\newcommand{\X}{X}
\newcommand{\Xg}{\X_G}
\newcommand{\Xfree}{\X_{\text{free}}}

\newcommand{\x}{x}

\renewcommand{\xi}{\x_{I}}
\renewcommand{\path}{p}
\newcommand{\pathstar}{\path^{\star}}
\newcommand{\Rpos}{\R_{\geq 0}}

\def\dof{\textsc{dof}\xspace}

\definecolor{crrtstar}{HTML}{90EE90}
\definecolor{cbitstar}{HTML}{ADD8E6}
\definecolor{cprm}{HTML}{800080}
\definecolor{cprmstar}{HTML}{40E0D0}
\definecolor{crrtconnect}{HTML}{FF6347}
\definecolor{cest}{HTML}{FFA500}
\definecolor{csst}{HTML}{DDA0DD}
\definecolor{cfmt}{HTML}{000000}
\definecolor{ckrrt}{HTML}{00FFFF}
\definecolor{ckpiece}{HTML}{008000}
\definecolor{cait}{HTML}{FFFF00}
\definecolor{ctrrt}{HTML}{DA70D6}
\definecolor{clbtrrt}{HTML}{FFD700}

\newcommand{\sqboxs}{1.5ex}
\newcommand{\sqbox}[1]{\textcolor{#1}{\rule{\sqboxs}{\sqboxs}}}

\newcommand\addSubCaption[1]{
    \caption{\centering\scriptsize
#1}
    \vspace*{-0.25cm}
}

%% file: src/00_abstract.tex
\begin{abstract}
    Sampling-based motion planning is one of the fundamental paradigms to generate robot motions, and a cornerstone of robotics research. 
    This comparative review provides an up-to-date guideline and reference manual for the use of sampling-based motion planning algorithms. 
    This includes a history of motion planning, an overview about the most successful planners, and a discussion on their properties. It is also shown how planners can handle special cases and how extensions of motion planning can be accommodated.
    To put sampling-based motion planning into a larger context, a discussion of alternative motion generation frameworks is presented which highlights their respective differences to sampling-based motion planning. 
    Finally, a set of sampling-based motion planners are compared on $24$ challenging planning problems. 
    This evaluation gives insights into which planners perform well in which situations and where future research would be required. 
    This comparative review thereby provides not only a useful reference manual for researchers in the field, but also a guideline for practitioners to make informed algorithmic decisions.
\end{abstract}

%% file: src/01_introduction.tex
\section{Introduction}

The last decades have seen remarkable progress in the field of robotics. An area of growth has been the development of sampling-based motion planning methods~\cite{Kavraki1996, Hsu1999, Kuffner2000, Karaman2011}, which have enabled applications such as robotics construction~\cite{Hartmann2021TRO}, multi-robot coordination~\cite{hoenig_2018}, autonomous driving~\cite{claussmann2019review}, industrial manufacturing~\cite{murray2016robot}, and protein folding~\cite{Albluwi2012Molecular}. 

Sampling-based motion planning (SBMP) is an approach to the problem of finding a motion for a robot to move from A to B. The distinguishing characteristic of SBMP is that it relies on sampling configurations (placements of the robot) in order to quickly find feasible and, in certain cases, optimal robot motions. Studies~\cite{Chamzas2022BenchMaker} have shown this to be one of the most efficient ways to solve problems with high numbers of degrees of freedom.

However, despite this success, incoming researchers and practitioners have currently no up-to-date guideline to use SBMP methods. In particular, there is currently no historical treatise recapping the development of the field until the present day. There exists also no categorization of planners summarizing the developments of the last decade. Moreover, it is often unclear how SBMP methods differ from alternative motion generation frameworks such as motion optimization, motion primitives, search-based planning or control-based planning. Finally, it is often unclear which planner class is best suited for a particular application area, and which planner performs best for a particular scenario. 

To mitigate those gaps, this comparative review makes four core contributions:

\begin{enumerate}
    \item an overview and brief history of SBMP methods, which spans the years 1979 to 2023,
    
    \item an analysis and categorization of SBMP algorithms to give insights into different planners classes and extensions,

    \item a discussion of how SBMP compares to alternative motion generation frameworks (e.g.,~\cite{cohen2010search, Toussaint2017, cheng2021rmpflow, kober2013reinforcement}), and its advantages and disadvantages,
    
    \item a large-scale comparative evaluation on $24$ scenarios, comparing several SBMP methods in terms of success, runtime, and optimality.
\end{enumerate}

Those contributions make this comparative review an extensive guideline and reference manual to leverage the power of SBMP methods.

%% file: src/03_history.tex
\section{Motion Planning History and the Emergence of Sampling-based Methods}

The history of motion planning begins around 1979~\cite{Lozano1979} and continues as an ever-growing research field into the present. This 40-year period can be divided into four eras. First, the pre-sampling era, where fundamental results were discovered and the problem was rigorously analyzed. Second, the sampling-advent era, where the first planners based on random sampling were discovered. Third, the sampling-consolidation era, where many improvements on sampling-based planners were made. Fourth, the optimality and learning era, where the first asymptotically-optimal planners were discovered and where learning algorithms became a focus of the community.

\subsection{Pre-sampling Era (1979-1989)}

The start of the motion planning era can be placed around 1979, when Lozano-P\'{e}rez~\cite{Lozano1979} introduced the concept of the configuration space as a general framework to plan motions for arbitrary kinematic systems. This idea of planning through a configuration space \cite{Lozano1983} put the research field on a solid foundation and helped properly articulate the problem of finding a path through the configuration space as the motion planning problem, sometimes also called the piano mover's problem \cite{Schwartz1983pianoII}. 

After the motion planning problem was articulated, many researchers focused on the topic of computational complexity~\cite{Schwartz1983pianoII, Reif1979, Canny1987, Canny1988complexity}. Fundamental results of that time were the proof that motion planning in configuration space is \textsc{NP}-hard~\cite{Canny1987, Canny1988complexity} and the development of the celebrated Canny algorithm which solves the problem in single exponential time~\cite{Canny1988complexity}. While the proposed motion planning algorithms have strong guarantees, they were not suited for practical, relevant problems due to their high computational complexity. 

This was partially overcome by Khatib~\cite{Khatib1986}, who introduced the artificial potential-fields. This method uses attractive and repulsive artificial potential fields to drive the robot towards a desired goal while avoiding obstacles. Potential-fields were a popular method to control robots, with several extensions like the Laplacian potential field method~\cite{Sato1992}, addition of circular directions~\cite{Koditschek1987}, navigation functions~\cite{koditschek1990robot}, and numerical potential fields~\cite{Barraquand1992}.
While all those methods had tremendous influence on later algorithms like task-space control, they do sacrifice guarantees like completeness or optimality of the solution. 

\subsection{Sampling-advent (1990-1999)}

The second era of motion planning comprises the years 1990 until 2000, where sampling-based algorithms were first pioneered and showed remarkable results in terms of efficiency of computation~\cite{Latombe1999}. This era roughly starts with the works of Barraquand~\cite{Barraquand1991}, who improved the potential field approach using a Monte-Carlo random walking method which were instrumental for subsequent sampling-based planners. Sampling-based planners differed from other approaches by estimating the connectivity of the free configuration space through sampling. Despite not having an explicit representation of the configuration space, several of them were able to hold the property of probabilistic completeness, meaning that they will find a solution when time goes to infinity. 

The most popular sampling-based approaches came in two categories. The first category were graph-based planners, like the probabilistic roadmap planner (PRM)~\cite{Kavraki1996}, which randomly samples configurations, connects them to a graph, and uses targeted sampling to connect graphs. Several planning queries can then be answered using the same graph. The second category were tree-based planners, like the expansive-space trees (EST)~\cite{Hsu1999, Sanchez2003} and the rapidly-exploring random tree (RRT) algorithm~\cite{Lavalle1998, Kuffner2000}. They both grow a tree from a start configuration by randomly sampling configurations and connecting them to the tree till the goal is reached. The growth of sampling-based methods was starting~\cite{barraquand1996random}.

\subsection{Sampling-consolidation (2000-2009)}

After several sampling-based planners were developed, it became clear that there are several areas where improvements could lead to another leap in terms of runtime. One of those areas were biased sampling functions, which do not sample uniformly the space, but bias samples towards certain areas. Several approaches were developed during this time, like sampling near or on the surface of obstacles~\cite{Amato1998}, sampling inside narrow passages~\cite{Hsu1998, hsu_2003}, Gaussian sampling around current frontier states and obstacles~\cite{boor_1999}, sampling restricted to workspace geometries~\cite{VanDenBerg2005} and workspace decompositions~\cite{Yang2005, kurniawati2004}, sampling on the medial axis of the environment~\cite{wilmarth_1999}, utility-based sampling to connect separate regions of roadmaps to each other~\cite{burns_2005}, and sampling in areas that are deemed difficult~\cite{Kavraki1996}. The dynamic-domain RRT~\cite{Yershova2005, Yershova2009} extended tree nodes based on their estimated exploration ability. 

\subsection{Optimality and Learning period (2010-today)}

Before the year 2010, sampling-based planning algorithms did not explicitly consider optimality. Instead, optimality was supposed to be a post-processing step, where a sampling-based planner provides a path which is then fed to an optimizer~\cite{geraerts2007creating} for further improvement~\cite{VanDenBerg2008,raveh2011little,luna2013anytime}. In 2010, another breakthrough occured, where the star versions of PRM and RRT, PRM* and RRT*~\cite{Karaman2011, solovey2020critical}, were developed. PRM* and RRT* are guaranteed to be asymptotically optimal, meaning they converge to the optimal solution in terms of path length at the limit, as the number of samples goes to infinity.

Besides achieving optimality guarantees, another important step was the integration of machine learning into motion planning. Robots often operate in similar environments solving similar motion planning problems. This motivated the use of past planning experiences to expedite the search in future problems. One approach of leveraging past experiences included retrieving a past solution that is similar to the current problem and repairing it. Past solutions are stored either in the form of path libraries  \cite{Berenson2012lightning}, sparse roadmaps \cite{Coleman2015Thunder}, or local obstacle roadmaps \cite{Lien2009}. Other approaches learn sampling distributions that can be used to bias the search of sampling based planners. Some methods learn sampling-distributions that are problem invariant \cite{Lehner2018} conditioned on the workspace description \cite{Chamzas2021}. More recent approaches based on deep learning can learn sampling distributions from past examples conditioned on workspace information, start, and goal information \cite{Ichter2018, Chamzas2022}. A recent review on learning for sampling-based planners summarizes those works~\cite{McMahon2022learningreview}.  



%% file: src/04_algorithms.tex
\section{Motion Planning\label{sec:motionplanning}}

The goal of motion planning is to develop algorithms to move mechanical systems (robots) from a start state to a goal region~\cite{choset2005principles, lavalle_2006, lynch2017modern}. 
A mechanical system consists of links and joints which can exist in different configurations or states based on the position or velocity of their joints.  
The set of all states of a system is called the \emph{state space}\footnote{State space is used here as a general umbrella term to denote all spaces like configuration space, joint space, Cartesian space, parameter space, or phase space.}. 
The state-space is denoted by the letter $\X$ and its elements as $x$.
\begin{marginnote}[]
\entry{State space}{The set of all states uniquely describing a system plus additional structure like metrics, constraints, dynamics, or topology.}
\end{marginnote}

Not all states in the state space are physically feasible---they might violate a \emph{constraint} (see Sec.~\ref{sec:constraint_functions}). Constraints divide the state space $\X$ into the constraint-free region $\Xfree$ and its complement $\X \setminus \Xfree$.
A \emph{motion planning problem} is a tuple $(\Xfree, \xi, \Xg)$, representing the task of finding a path, a continuous function $\path: [0,1] \rightarrow \Xfree$, from a start state $\xi \in \Xfree$ to a goal region $\Xg \subseteq \Xfree$. The set of all feasible paths $P$ is defined as the \emph{path space} $P(\Xfree, \xi, \Xg)$.


Motion planning problems can have several variations. The most important ones are
\begin{itemize}
    \item \textbf{Path planning.} The geometrical problem ignoring the velocity, time, or dynamics of the system~\cite{lynch2017modern}. This is often referred to as the piano mover's problem~\cite{Schwartz1983pianoII}.
    \item \textbf{Kinodynamic planning.} Planning with a dynamical system and possible constraints on velocity, acceleration, or torque. A kinodynamic planning problem can be defined as a tuple $(\Xfree, \xi, \Xg, f)$, where $f$ are the dynamical equations~(Sec.~\ref{kinodynamic_planning_extension}).
    \item \textbf{Optimal planning.} The problem of finding a global optimal path. An optimal path is a path $\path$ which minimizes a given cost functional $c: P \rightarrow \Rpos$. An \emph{optimal motion planning problem} is a tuple $(\Xfree, \xi, \Xg, c)$, where the goal is to find a feasible path $\pathstar$, such that $c(\pathstar) = c^{*}$ and $c^{*}$ is the minimum cost over the path space $P$.
\end{itemize}

The discussion that follows will focus on path planning, while kinodynamic planning is discussed in Sec.~\ref{kinodynamic_planning_extension}. Optimal planning is interleaved with the description of path planning and kinodynamic planning.

\subsection{State Space Structure\label{sec:algorithms:assumptions}}

The state space $\X$ needs to have additional structures. This includes it being a topological space~\cite{farber2003topological}, which is required to define the notion of a path and of path-connectedness. Most planners further assume that the state space is a \emph{manifold}. A manifold is a topological space, which locally resembles an Euclidean space $\R^n$. This is an important assumption, because mechanical systems are naturally modelled by manifolds~\cite{lee_2003}. Additional structures also include metric functions and constraints.

\subsection{Metric Function\label{sec:metric_space}}

Most planners require a way to measure distances. This can be achieved by adding a \emph{metric function} defined as $d: \X \times \X \rightarrow \Rpos$. Given any elements $x,y,z \in \X$, a metric function is characterized by the following three assumptions:

\begin{enumerate}[label=M.\arabic*,ref=M.\arabic*]
    \item $d(x,y)=0 \Longleftrightarrow x=y$ (Identity of indiscernibles),\label{M1}
    \item $d(x,y)=d(y,x)$ (Symmetry),\label{M2}
    \item $d(x,y) \leq d(x,z) + d(z,y)$ (Triangle inequality).\label{M3}
\end{enumerate}

If a metric cannot be defined on a problem, less-restrictive functions can be defined, which sacrifice one of the assumptions of a metric. The most important ones are:

\begin{itemize}
    \item \textbf{Pseudometric}: Replace \ref{M1} by the assumption $d(x,x) = 0$ for all $x \in \X$~\cite{Cech1966Pseudometric}. An example is end-effector distance, which is zero for different inverse kinematic solutions.
    \item \textbf{Quasimetric}: Remove assumption on symmetry \ref{M2}~\cite{wilson1931quasi}. An example are time-dependent state spaces, where the robot can only move forward in time~\cite{grothe2022ICRA}.
    \item \textbf{Semimetric}: A metric invalidating the triangle inequality \ref{M3}~\cite{wilson1931semi}. A simple example are measurement errors violating the triangle inequality~\cite{wang2007towards}. 
\end{itemize}

However, one has to be careful sacrificing metric properties, because some methods like nearest neighbor computations can depend on them. There exists also sampling-based planners that operate without a metric~\cite{Ladd2004}. 

\subsection{Constraint Functions\label{sec:constraint_functions}}

A constraint function codifies what a state has to satisfy to be considered feasible. 
Feasibility is problem-dependent and often involves avoiding obstacles, pushing an object, turning an object, or requiring that the robot stays close to a surface. 
Those tasks can be formalized using a constraint function $\phi: \X \rightarrow \R$, which evaluates to less or equal to zero if a state is satisfied (or constraint-free), or to a value larger than zero otherwise. Most classical motion planning problems can be formulated using one of the following constraints.

\begin{itemize}
    \item Collision constraint. Reject all states where robot links are in collision, either to other robot links (self-collisions), or to links in the environment. The constraint function often only returns binary values, but can also be extended to return clearance or penetration depth.
    \item Joint limit constraint. Reject all states which violate joint limits on the robot. This might be a simple interval check for some actuators, but might also involve coupled joint checks where cables restrict the motion.
    \item Tool-center-point (Tcp) constraint. Reject states where the Tcp of the robot is outside certain limits. The Tcp is usually a predefined coordinate frame on the robot, where a tool is attached e.g., for spraying, painting, or welding. 
    \item Kinodynamic constraint. Reject states which do not fulfil velocity, acceleration, or jerk limits on the robots joints. This is important for dynamical systems, where not all geometric paths have a valid velocity profile.
\end{itemize}

\section{Sampling-based Motion Planning}



Sampling-based motion planning is the idea of implicitly representing the state space through the use of a sampling function (Sec.~\ref{sec:samplingfunction}). The sampling function generates a sequence of states which can be connected using a local planner (Sec.~\ref{sec:steeringfunction}). 
To coordinate sampling and local planning, different planners have been developed, which are categorized in Sec.~\ref{sec:plannercategorization} either as tree-based planners (often used in single query problems), or graph-based planners (often used in multi-query problems).
Over the years several general-purpose improvements have been proposed~(Sec.\ref{susbec:general_improvements}) that enhance different planner aspects. 
Besides performance improvements, there are many special-purpose sampling-based planners addressing variations of the canonical motion planning problem such as planning in unbounded spaces (Sec. \ref{subsec:unbounded}), infeasible problems (Sec. \ref{subsec:infeasible}), planning with kinodynamic constraints (Sec.\ref{kinodynamic_planning_extension}), and other motion planning extensions (Sec. \ref{subsec:extentions}). 

\subsection{Sampling Function\label{sec:samplingfunction}}

A sampling function generates an infinite sequence of elements of the state space as $S = \{s_1,s_2,\cdots\}$. For planners to offer guarantees, the sequence $S$ is required to be dense in the state space $\X$, which means that every point of $X$ is arbitrarily close to a member of $S$.

Sampling functions are classified as unbiased or biased methods. An unbiased method, or uniform sampling, draws elements of the state space, whereby each outcome has an equal chance~\cite{bertsekas2008introduction}. Biased methods, however, change the probability distribution by biasing sampling towards interesting regions of the state space. While different biases can be used, sampling-based systems often favor one of the following three.

A first method is obstacle-based sampling~\cite{Amato1998}. States close to an obstacle have a higher chance of being selected, i.e., there is a bias towards the boundary of the free state space $\Xfree$. Well known obstacle-based samplers are the Gaussian sampling method~\cite{boor_1999}, and the bridge-based sampling method~\cite{hsu_2003}. This bias often improves planning in narrow passages~\cite{Amato1998}, but also imposes an implicit bias on path length, which might interfere with other cost functionals like clearance. 

A second method is clearance-based sampling~\cite{wilmarth_1999}. Those samplers prioritize samples which increase clearance, i.e., the distance between robot and environment. This can be achieved by sampling a feasible state, and making random steps to improve its clearance~\cite{Verginis2022kdf}. Clearance-based sampling can mitigate execution uncertainty on real robots, but computing clearance queries is often expensive.

A third method is deterministic sampling~\cite{janson2018deterministic, palmieri2019dispertio}. Deterministic samplers reproduce the same sampling sequence in each run. Those sequences can be learned from similar environments to bias samples towards optimal paths~\cite{Ichter2018}, or they can achieve better distributions by minimizing the largest uncovered area (low-dispersion). Examples of low dispersion sequences are Halton sequences and Sukharev grids~\cite{lavalle_2006}. 


\subsection{Local Planning\label{sec:steeringfunction}}
%

To find a continuous connected path, it is necessary to connect two state-space samples to each other and return a path segment (an edge) connecting them. This is accomplished using a \emph{local planner}. A local planner is called local because the path segment often connects samples over a short distance using simple path segments (e.g., a straight line).
It is not global in the sense that it does not address the complete global motion planning problem. A local planner is usually fast, but the produced path might not satisfy the constraints of the problem e.g., it might be in collision.

Depending on the category of the motion planning problem (Sec.~\ref{sec:motionplanning}) the local planner might need to satisfy additional constraints.
In the case of optimal planning the produced path must be a lower-bound on the true solution cost. 
In the case of kinodynamic motion planning the differential constraints must be satisfied (see Sec.~\ref{kinodynamic_planning_extension}). Having efficient and optimal local planners is an active area of research with different methods leveraging local optimization \cite{choudhury2016regionally} or learning \cite{faust2018prm} to improve their performance.

\subsection{Categorization of Sampling-based Planners\label{sec:plannercategorization}}
\input{images/figure_basic_prm}
Sampling-based planners can be classified into two main categories~\cite{choset2005principles, lavalle_2006, lynch2017modern}, \emph{graph-based} (or roadmap-based) and \emph{tree-based} planners. 
Algorithm~\ref{alg:prm} includes the \textsc{basic-prm}~\cite{Kavraki1996} as a representative example of graph-based planners.
A graph-based planner produces a graph by sampling (Sec.~\ref{sec:samplingfunction}) constraint-free states and adding them to the graph \mbox{(Line 4-5)}. 
Edges to nearest neighbors are added using a local planner (Sec.~\ref{sec:steeringfunction}) \mbox{(Line 6-10)} to connect to the graph. The planner terminates if a planner terminate condition (PTC)\footnote{The planner terminate condition can for example be a successful solution, a timeout, or a maximum number of iterations.} is fulfilled \mbox{(Line 3)}.
These planners are often called \emph{multi-query}, because the graph can be reused for changing start states or goal regions. 
Examples include the probabilistic roadmap planner (PRM)~\cite{Kavraki1996, Karaman2011}, and the sparse roadmap method~\cite{dobson_2014}.

A representative tree-based planner is \textsc{basic-rrt} (Algorithm~\ref{alg:rrt})~\cite{Kuffner2000}. 
Using a sampling function, a random sample from the state space is chosen \mbox{(Line 4)}. 
This random sample is used to extend from the nearest state in the tree \mbox{(Line 5-7)}, that is to generate a path starting from the nearest state to the random sample with a local planner. If the resulting edge is constraint-free, it is added to the tree \mbox{(Line 8-9)}. These planners are often called \emph{single-query}, because the trees need to be recomputed for different start states or goal regions. Examples include the rapidly-exploring random trees (RRT) planner~\cite{Kuffner2000, Karaman2011}, the expansive-space trees (EST) planner~\cite{Hsu1999}, the lower-bound trees RRT (LBT-RRT)~\cite{Salzman2013}, and the fast-marching trees planner (FMT)~\cite{Janson2015}.


\subsubsection{General-purpose Planner Improvements}\label{susbec:general_improvements}

Planner efficiency can improve by adding or improving one of the following components. Efficiency here can mean reducing memory footprint, decreasing runtime, decreasing number of samples, or improving path cost.

\textbf{Lazy Checking}. A lazy version of a planner~\cite{Sanchez2003, Bohlin2000, Hauser2015, Mandalika2019} ignores edge constraint checking during planning. This reduces memory footprint and speeds up runtime. Once a solution has been found, the edges are checked for constraint violations. In the case of a constraint violation, the edge is removed from the tree or graph and planning is continued. This method can be seen as a constraint relaxation method, where a simplified problem is solved first, and this information is leveraged to solve the original problem~\cite{Orthey2019}.

\textbf{Bidirectionality}. Most tree-based planners can be extended to plan not only with one, but two or more trees. An example is the bidirectional RRT planner~\cite{Kuffner2000} which alternates between extending two trees grown from start and goal, respectively. On each successful extension, a connection between the nearest states in the trees are tried. This approach can also be extended to optimal versions~\cite{Klemm2015}. In general, this decreases runtime significantly. 

\textbf{Sparsity}. Most planners can also be made sparse. A sparse planner often ignores samples which are inside of a visibility radius of a given node in the graph or tree~\cite{Simeon2000}. Another way to ensure sparsity is to only keep the best cost-to-come states in certain regions of the state space~\cite{Li2016}. This is a good way to reduce the memory footprint of the planner.

\textbf{Optimality}. In the pioneering work by Karaman and Frazzoli~\cite{Karaman2011}, the authors show that many planners can be adapted to make them converge to the global optimal solution (asymptotic optimality). This involves having an adaptive nearest neighbors radius for graph-based planners~\cite{solovey2020critical}, or using a tree-rewiring operation after each sampling iteration~\cite{Salzman2016}. This approach usually improves the cost of paths significantly.

\textbf{Admissible Heuristics}. Admissible heuristics are lower bound estimates on the cost to reach a goal~\cite{pearl1984heuristics}. One important category are \emph{informed sets}~\cite{Gammell2014}, which describe all states which can improve the solution quality. Examples include the batch-informed trees planner (BIT*)~\cite{Gammell2020}, and the advanced-informed trees planner (AIT*)~\cite{Strub2020Advanced}. Another category are constraint relaxations~\cite{Orthey2019}, where (multiple) levels of projections are used to simplify the problem, as in the KPIECE method~\cite{Sucan2009, Sucan2011}, the quotient-space rapidly-exploring random tree planner (QRRT*)~\cite{Orthey2019}, and the hierarchical fast marching tree planner (HFMT*)~\cite{Reid2019}. A good admissible heuristic can decrease the planner runtime significantly without sacrificing completeness or optimality.

\textbf{Parameter-Tuning}. Most planners have parameters e.g., the number of nearest neighbors K in \textsc{Basic-PRM} (Algorithm~\ref{alg:prm}), that need to be chosen before planning. In certain problem scenarios these parameters can significantly affect planning performance. Although planning frameworks such as \textsc{OMPL}~\cite{Sucan2012} often offer reasonable defaults, choosing an appropriate set of hyperparameters is considered an open research problem. Some recent works have investigated bayesian optimization to automatically choose these parameters~\cite{moll2021hyperplan, cano2018automatic}.

\subsubsection{Planner Properties}

Sampling-based planners set themselves apart from competing motion generation frameworks (see Sec.~\ref{sec:competingframeworks}) by providing desirable guarantees. One guarantee is \emph{probabilistic completeness}. Probabilistic completeness states that a planner will find a solution path if one exists, when time goes to infinity. 
\begin{marginnote}[]
\entry{Probabilistic Completeness}{The planner will find a solution if one exists, when time goes to infinity.}
\end{marginnote}

Another important guarantee is \emph{asymptotic optimality}~\cite{Karaman2011, Gammell2020Survey}. Asymptotic optimality states that a planner will find the global optimal solution path when time goes to infinity. Algorithms with this property, like RRT*, PRM*~\cite{Karaman2011}, or BIT*~\cite{Gammell2018}, are able to continuously improve the solution cost until they eventually converge to the global optimal solution~\cite{Karaman2011}. A slightly weaker notion is \emph{asymptotic near-optimality}. This is a variant of asymptotic optimality stating that a planner will find a solution when time goes to infinity whereby the solution cost is $\epsilon$-near to the cost of the optimal solution. Some planners like SPARS~\cite{dobson_2014} provide this weaker notion to trade-off memory consumption with optimality guarantees.
\begin{marginnote}[]
\entry{Asymptotic Optimality}{The planner will find the global optimal solution when time goes to infinity.}
\end{marginnote}

\subsection{Special Cases of Motion Planning}
\input{images/figure_unbounded_space}

While several sampling-based planners are general enough to solve any motion planning problem, there are some special cases, which require special care.

\subsubsection{Unbounded Space}\label{subsec:unbounded}

Some problems, like planning in space-time, require sampling of an unbounded state space (see Fig.~\ref{fig:spacetime}). 
However, most sampling sequences require a bounded space to generate dense samples. 
To resolve this discrepancy, one option is to use adaptive goal regions~\cite{grothe2022ICRA}, where lower and upper bounds are shifted to keep asymptotic optimality while having a bounded region to enable sampling. 
Another option is to select existing nodes using a selection-extension scheme, where nodes are selected at random or based on their utility for expansion~\cite{Yershova2005}. 
Those nodes can be extended into random directions or via dynamics-driven propagation functions~\cite{Li2016}.

\input{images/figure_infeasible_problem}

\subsubsection{Infeasible Problems}\label{subsec:infeasible}

While sampling-based planners usually provide probabilistic completeness guarantees, they often cannot deal with infeasible planning problems (see Fig.~\ref{fig:infeasibleproblem}). 
To handle infeasible problems, there are different methods which can be applied. 
One method is based on sparse roadmaps~\cite{Simeon2000, dobson_2014}. 
A sparse roadmap has a given visibility radius, such that new samples inside the visibility radius are rejected, and the number of samples added to the roadmap goes to zero if time goes to infinity. 
If no samples can be added for a certain period, a probabilistic estimate of infeasibility can be given~\cite{Simeon2000, Orthey2021ICRA}. 

Another method is based on infeasibility certificates~\cite{Mccarthy2012, Varava2020, li2021learning}. 
Such methods can tackle lower-dimensional problems, where samples are used to create a closed hull around the start state or around the goal region to verify that a problem is infeasible~\cite{li2021learning}.


%% file: images/figure_basic_prm.tex
\begin{wrapfigure}{R}{0.5\textwidth}
    \begin{minipage}{0.5\textwidth}
    \vspace*{-1cm}
    \begin{algorithm}[H]
    \scriptsize
    \caption{Basic-PRM \label{alg:prm}}
    \begin{algorithmic}[1] 
    \Procedure{Basic-PRM}{$\xi $}  
        \State G.addNode($\xi$) 
        \While{PTC is false}
            \State $\x_{new} \gets $ valid sample from $\Xfree$
           \State G.addNode($\x_{new}$) 
           \State $\mathcal{N}(x_{new}) \gets $ K closest neighbors of $x_{new}$
           \For{each $x_{near} \in \mathcal{N}(x_{new})$} 
            \State $e \gets$ Local plan $x_{new}$ to $x_{near}$
            \If{$e \in \Xfree$ and $e \notin$ G.edges()}
            \State  G.addEdge($e$) 
            \EndIf 
            \EndFor 
        \EndWhile
    	\State \Return G
    \EndProcedure
    \end{algorithmic}
    \end{algorithm}
    \begin{algorithm}[H]
    	\scriptsize
    	\caption{Basic-RRT \label{alg:rrt}}
    	\begin{algorithmic}[1]
        \Procedure{Basic-RRT}{$\xi$}  
        \State T.addNode($\xi$)
            \While{PTC is false}
    		\State $\x_{rand} \gets $ sample from $\X$
    		\State $\x_{near} \gets $ nearest node in T to $\x_{rand}$ 
    		\State $\x_{new} \gets$ Extend $x_{near}$ towards $x_{rand}$
    		\State $e \gets$ Local plan $x_{new}$ to $x_{near}$
    		 \If{$e \in \Xfree$}
    		\State  T.addEdge($e$) 
    		\EndIf
            \EndWhile
        \State \Return T
        \EndProcedure
    \end{algorithmic}
    \end{algorithm}
    \end{minipage}
\end{wrapfigure}

%% file: images/figure_unbounded_space.tex
\begin{wrapfigure}{R}{0.3\textwidth}
\vspace*{-0.5cm}
        \centering
        \includegraphics[width=1.0\linewidth]{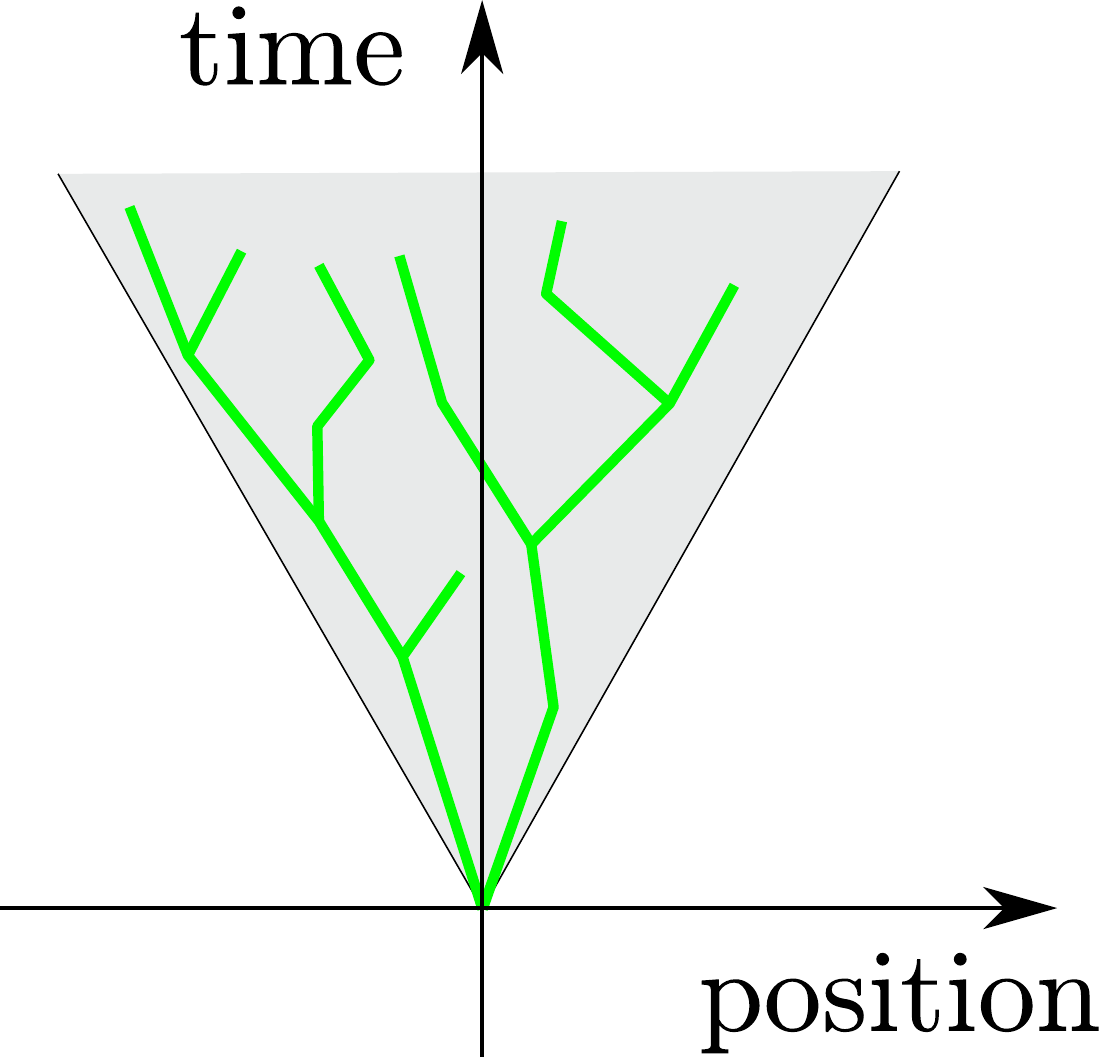} 
    \caption{\small Unbounded time-dependent
state space, with no natural time limit.\label{fig:spacetime}}
\vspace*{-0.9cm}
\end{wrapfigure}

%% file: images/figure_infeasible_problem.tex
\begin{wrapfigure}{R}{0.34\textwidth}
\vspace*{-0.2cm}
        \centering
        \includegraphics[width=1.0\linewidth]{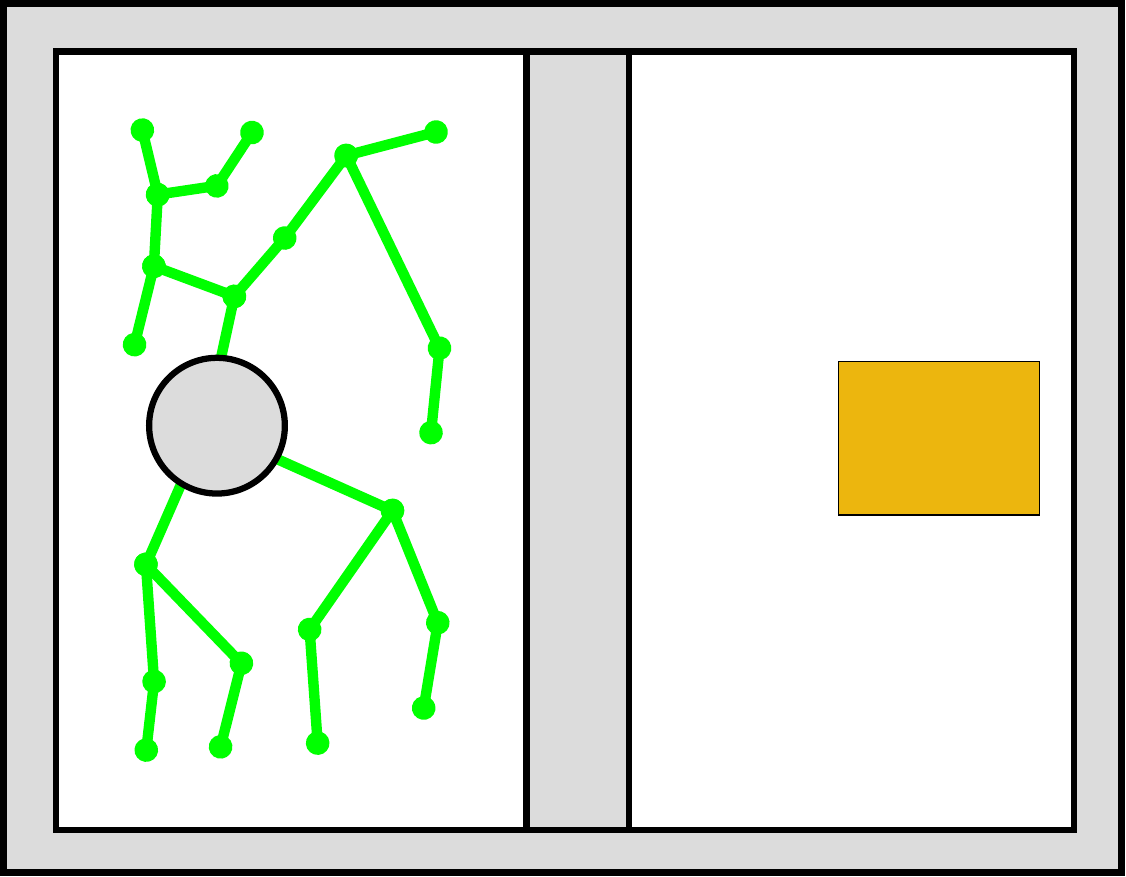} 
    \caption{\small Infeasible problem, where a disk robot (grey) can plan motions (green), but cannot reach a goal (orange) due to obstacles (grey).\label{fig:infeasibleproblem}}
    \vspace*{-0.3cm}
\end{wrapfigure}

%% file: src/05_extensions.tex
\subsection{Kinodynamic Motion Planning\label{kinodynamic_planning_extension}}

In addition to kinematic constraints, many realistic systems have to satisfy dynamics and constraints on velocity, acceleration, or torques. This is known as kinodynamic motion planning~\cite{lavalle_2006}. The dynamics of kinodynamic systems can be expressed as:
\begin{equation*}
    \dot{x} = f(x,u),
\end{equation*}
whereby $\dot{x}$ is the derivative of $x$, $u \in U$ the controls of the system, $U$ the space of applicable controls, and $f$ the dynamical systems equation. This dynamics equation can be seen as imposing a constraint on the allowable paths the system can take. There are two main ways of how planners can handle dynamics.


\textbf{Steering method}. In this approach, given an initial state $x_1$ and a target state $x_2$, the dynamics $f$ are used to analytically or numerically compute a control $u$ and time $t$ which respects the dynamics and moves the state from $x_1$ to $x_2$. This simplifies planning, but steering might be difficult and costly to compute. Examples of systems with analytical steering methods are Dubin's car~\cite{dubins1957curves} and the Reeds-Sheep car~\cite{reeds1990optimal}.
In general, a two-point boundary value problem (BVP) between two states needs to be solved. The solution of a BVP provides a local planner which can be used in any geometric planning algorithm to eventually solve the kinodynamic problem. 

\textbf{Forward propagation} The second approach uses forward propagation, whereby an initial state $x_1$ and a control $u$ is used, and the system is forward propagated for a specified time $t$ to compute the next state $x_2$. This is advantageous since the dynamics can be treated as a black box. Kinodynamic planners like Kinodynamic-RRT~\cite{Kuffner2000} can solve those problems by using random control sampling to propagate states forward. Asymptotic optimality can be achieved using the meta planner AO-RRT~\cite{hauser2016asymptotically} which relies on the Kinodynamic RRT to produce feasible solutions in a combined cost-state space. A kinodynamic planner which combines asymptotic near-optimality and efficiency is the stable sparse RRT (SST)~\cite{Li2016}, which grows a tree in state space while using pruning to maintain only the locally best solutions~\cite{Verginis2022kdf}. An asymptotic optimal version (SST*) can be achieved by iteratively decreasing the pruning resolution~\cite{Li2016}.

\subsection{Extensions Solved by Sampling-based Motion Planning}\label{subsec:extentions}

An extension of motion planning is defined as a problem which imposes additional structures onto the state space. This additional structure can either be dictated by the problem itself, such as contact constraints, or partial-observability. Or the structure is added intentionally to improve planner performance, like projections, or differentiability. This section gives a non-exhaustive list of additional structures, and how sampling-based planners can be extended to accommodate them.

\subsubsection{Projection-based Motion Planning}

Many high-dimensional problems can often be simplified by introducing projections~\cite{Orthey2019, Sucan2011, Reid2019}. Projections allow 
planning over different levels of abstraction. 
There are different ways to plan with projections, either by using them as biased samplers~\cite{Sucan2011}, or by using projections to adjust and guide the sampling. If projections are admissible, i.e., solving a simplified problem is a necessary conditions on the original problem, properties like asymptotic optimality can be guaranteed~\cite{Orthey2019, Reid2019}.
While projections usually speed up planning significantly~\cite{Orthey2019}, it is often difficult to define them for a new problem domain. 

\subsubsection{Differentiable Motion Planning}

Most functions used in motion planning, like costs, goals, or constraints, are often differentiable. Exploiting those functions is studied under the topic of differentiable motion planning~\cite{mukadam2018IJRR}. While most sampling-based planner avoid differentiable information, there is evidence that differentiable information can be useful to converge faster to optimal solutions~\cite{Kamat2022IROS}. Differentiable planners need to carefully weigh when and if computing differentiable information is useful. Recent work in this direction combines optimization with sampling by adding graph optimization~\cite{alwala2021joint}, improving approximately valid paths~\cite{Kamat2022IROS}, or optimizing the steering function~\cite{choudhury2016regionally}.

\subsubsection{Planning with Manifold Constraints}

Most constraints in classical motion planning can readily be sampled to construct satisfiable state space elements. However, complex constraints, like contacts between robot and environment, often introduce regions in the state space having zero measure, which have zero chance to be sampled. Dealing with such constraints is called motion planning with manifold constraints~\cite{kingston2018sampling}. There are different ways of how to deal with those constraints. For example, the volume of the constraints can be relaxed to simplify the problem, or random samples can be projected back to the constraints~\cite{kingston2018sampling, berenson2011task}. 

\subsubsection{Motion planning in Dynamic Environments} 

Often, robots have to deal with obstacles which might move, appear, or disappear. Dealing with such obstacles is studied under the topic of \emph{planning in dynamic environments}~\cite{phillips2011sipp}. 
Since successful plans might get invalidated, specialized planners are required. Examples include RRTX~\cite{Otte2016}, which updates a goal-centered search tree on the fly, and precomputed roadmaps to quickly recompute solution paths~\cite{murray2016robot}.

\subsubsection{Belief Space Planning} 
Many realistic robots do not have access to a fully observable environment, but need to discover the world using sensors. This leads to the problem of \emph{planning in partially-observable environments}~\cite{kurniawati2022partially}.
To solve those scenarios the belief-space can be sampled~\cite{Phiquepal2022RAL}, which contains hypotheses about the world plus robot states. Since the space of hypotheses might get large, it is important to properly simplify those spaces to make them searchable~\cite{elimelech2022simplified}. 

\subsubsection{Planning in Force Fields}

Often, robots need to handle external forces arising from gravity, friction, or wind. Sampling-based motion planning frameworks can integrate forces as vector fields on the state space of the robot~\cite{ko2014randomized}. Planning methods in this area mostly focus either on handling wind disturbances in Unmanned Aerial vehicles (UAV)~\cite{tang2020vector}, or handling water draft in Autonomous Underwater Vehicles (AUV)~\cite{hernandez2019online, vidal2019online}.

%% file: src/06_competitions.tex
\section{Competing Motion Generation Frameworks\label{sec:competingframeworks}}

Sampling-based motion planning is a framework to generate motions for arbitrary mechanical systems. However, there are competing frameworks to generate motions. Those frameworks are briefly surveyed and advantages and disadvantages are listed relative to sampling-based motion planning.

\subsection{Motion Optimization}
 
A framework with complementary strengths to sampling-based planning is motion optimization. The idea behind motion-optimization is to formulate motion generation as an optimization problem~\cite{boyd2004convex}, and often focus on improving an existing path.

A distinction can be made between gradient-free and gradient-based methods. Gradient-free methods include shortcutting~\cite{geraerts2007creating} or hybridizing\footnote{Hybridizing refers to the operation of combining a set of (possibly high-cost) input paths to generate a single (low-cost) output path.}~\cite{raveh2011little} to improve path length. Interpolation of splines~\cite{Hauser2014} can be used to improve path smoothness.
Gradient-based methods often make the cost, the goal, and the constraints differentiable, which allows them to quickly find low-cost paths. Examples include general-purpose optimization frameworks like CHOMP~\cite{zucker2013chomp}, TrajOpt~\cite{schulman2014motion}  KOMO~\cite{Toussaint2017}, GPMP\cite{mukadam2018IJRR}, to optimize motions conditioned on arbitrary task requirements.

\textbf{Advantages.} Since most optimization-based methods use second-order information, they can quickly converge to low-cost solutions. Most algorithms can also handle information to push robots out of collisions, which makes those methods applicable even when a returned path is in collision~\cite{zucker2013chomp}.

\textbf{Disadvantages.} Optimization-based methods usually can only find locally optimal solutions, and lack guarantees on completeness or optimality. If the starting path is not constraint-free, one may end up with an invalid path---even if a feasible solution exists. 

\subsection{Motion Primitives}

Motion primitives~\cite{schaal2006dynamic, Kober2011ML, Kober2013IJRR, paraschos2018using} are predefined or learned motions to accomplish a certain task. They can be defined as dynamical systems in the state space, which provide a vector field along which the robot can move to reach a target, fulfill a task, or avoid an obstacle. Those primitives are loosely based on insights from neuroscience~\cite{giszter2015motor}, which have shown that animals are often moving by composing sets of motion primitives for reaching or walking motions~\cite{graziano2008intelligent}. 

To equip robots with motion primitives, they can either be learned or predefined based on task requirements like avoiding obstacles or reaching a target pose. 
Motion primitive frameworks like Riemannian motion policies~\cite{cheng2021rmpflow}, are able to compose complex motions by combining several simple task policies, whereby a policy is a function telling the robot where to move from anywhere in the state space.

\textbf{Advantages.} Policies can be used for reactive planning, where obstacles can be avoided in realtime~\cite{bhardwaj2022storm}. This can be a good choice for problems where feedback is crucial, like playing table tennis~\cite{Kober2013IJRR}.

\textbf{Disadvantages.} Motion primitives usually do not provide guarantees on completeness or optimality. Finding primitives is often difficult and task-dependent. To make the methods work, it is often necessary to fine-tune the primitives and control policies.

\subsection{Search-based Planning}

Search-based planning~\cite{koenig2004lifelong, cohen2010search, ren2022multi} differs from sampling-based motion planning by imposing a grid onto the state space. 
By connecting neighboring states in this grid, a state space graph can be constructed. 
Based on this graph, search-based planners like A*~\cite{hart1968formal} can find optimal motions with respect to the resolution of the grid. This is an efficient method to quickly find solutions in lower-dimensional state spaces~\cite{liu2018search}.

\textbf{Advantages.} Variants of A*-like algorithms can directly be used, which are guaranteed to solve a problem to the optimal solution---w.r.t. the resolution of the graph. Open-source software is available in the search based planning library (SBPL)\footnote{\url{https://github.com/sbpl/sbpl}}.

\textbf{Disadvantages.} Only resolution-completeness can be guaranteed depending on the grid resolution. If the resolution is too fine, planning time might be exceptionally expensive. If the resolution is too coarse, narrow passages cannot be found or traversed. Search grids are also difficult to employ in high-dimensions due to the curse of dimensionality.

\subsection{Control-based Planning}
Control algorithms are useful to drive a robot towards a desired goal state. An example is the Proportional–integral–derivative (PID) controller~\cite{araki2010control}, which uses current state information to compute the next input to the system. This can be combined with repulsive forces to push the system away from obstacle regions, as in the artificial potential fields approach~\cite{Khatib1986}. More global approaches are the Linear–quadratic regulator (LQR) method~\cite{klemm2020lqr} which optimizes a (locally) optimal path, and the Model predictive control (MPC) method~\cite{grandia2019feedback} which can find optimal path segments over a receding horizon. Controllers can often also be chained together to provide some form of guarantee, as in the funnels approach~\cite{majumdar2017funnel}, where multiple local controllers cover the state space. 

Controllers can also be learned using Reinforcement Learning~\cite{sutton2018reinforcement} approaches, where the learned controller (policy) tries to maximize the given reward signal.  
A reinforcement learning (RL) algorithm improves itself by updating an underlying value function, which is a measure for how desirable states are depending on the reward signal. 
Over time, learning algorithms like Q-learning~\cite{watkins1989learning} will eventually converge to the optimal policy.  For long-horizon plans, controllers are often combined with planning methods from the previous sections which compute a reference trajectory which is subsequently given to the controller.

\textbf{Advantages.} The above approaches achieve fast computation of locally optimal paths, whereby most controllers are reactive and simple to implement. Additionally, a controller can can continuously incorporate execution feedback.

\input{images/figure_success_cost_graph}
\textbf{Disadvantages.} Controllers might get stuck in local minima, or return a sub-optimal path. There is often no guarantee on either completeness or optimality. If used inside a planning framework, computation time might become the bottleneck.

%
%
%

%% file: images/figure_success_cost_graph.tex
\begin{wrapfigure}{r}{0.5\textwidth}
\vspace{-1cm}
  \begin{center}
    \includegraphics[width=0.45\textwidth]{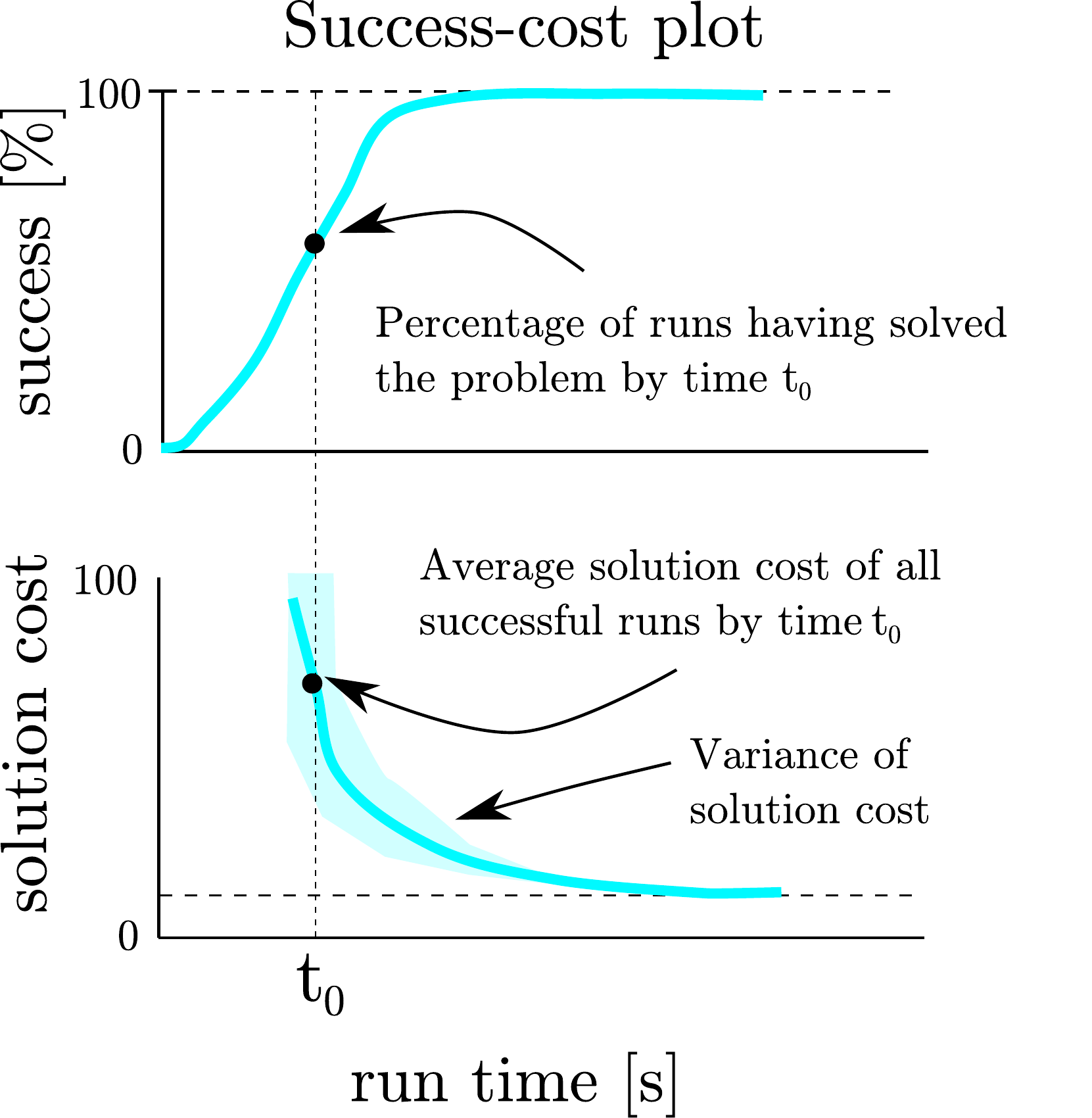}
  \end{center}
  \caption{\small{Success-cost plot explanation.}\label{fig:successcost}}
\vspace{-2cm}
\end{wrapfigure}

%% file: src/07_evaluations.tex
\section{Comparative Evaluations}

To reveal the relative performance of planning algorithms, several motion planners from the open motion planning library (OMPL)\footnote{\url{https://github.com/ompl/ompl} and \url{https://ompl.kavrakilab.org}}~\cite{Sucan2012, Moll2015} are evaluated on a set of $24$ scenarios. 
For each scenario, a set of applicable planners is selected and run for $100$ runs with a scenario-dependent timeout up to $300$s. 
For all experiments the default OMPL parameters of the planners were used. 
Changing those parameters could potentially influence the results~\cite{Chamzas2022BenchMaker}, but the size of the experimental dataset prevents fine tuning.
As hardware, a $4$-core, $8$GB RAM laptop running Ubuntu $16.04$ is used. 
The results are shown as success-cost plots (see Fig.~\ref{fig:successcost}), with time on the $x$-axis, success rate and solution cost\footnote{Solution cost is in all cases path length if not explicitly mentioned.} on the $y$-axis. 
In the case of a non-optimizing planner, the average cost of the first solution is displayed as a single cross. When a color is not displayed, this means that the planner was not able to find any solution paths.

\subsection{Classical Experiments}
\input{evaluations/Classical/classical_experiments}

This set of experiments includes classical motion planning problems, which frequently appear in the motion planning literature. In each scenario, the task is to move a rigid body from a start to a goal region. The scenarios are
\begin{enumerate}
   \item \textbf{Barriers}: A 3-\dof problem consisting of a horseshoe-like robot having to traverse a rectangle with several obstacles.
   \item \textbf{Maze}: A 3-\dof problem, where a small disk-like robot has to traverse a maze.
   \item \textbf{Random polygons}: A 3-\dof problem, where a a disk-like robot has to traverse a room with random polygonal obstacles.
   \item \textbf{Piano movers problem}: A 6-\dof problem, with a piano moving across a room.
   \item \textbf{Cubicles}: A 6-\dof problem with a floating L-shape which has to traverse a cubicle.
   \item \textbf{Apartment}: A 6-\dof problem, where a table moves through an apartment with furniture.
\end{enumerate}

Fig.~\ref{fig:experiments:classical} shows the results. It can be seen that BIT*, EST, FMT and RRT-Connect perform well in terms of success rate on 4 out of 6 scenarios. FMT is slightly slower than the other planners, but is the only one solving the Home scenario in $50\%$ of the cases. In terms of cost, BIT* has the best convergence in 5 out of 6 scenarios. Graph-based planners like PRM take longer to solve those problems. However, this performance is offset by the reusability of graphs for future planning queries. This evaluation underlines a common observation among practitioners, that there is no single planner that has the best performance across all scenarios. Choosing a suitable planner depends significantly on the robot and environment.

\subsection{Manipulation Experiments}
\input{evaluations/Manipulation/manipulation_experiments}

This set of experiments includes problems from the MotionBenchMaker dataset \cite{Chamzas2022}. In each problem the position of objects is varied relative to the robots.
\begin{enumerate}
   \item \textbf{Ur5 Small Shelf}: 6-\textsc{dof} problem where the \textsc{ur}5 robot starting outside a small shelf reaches inside the shelf to pick up a cylindrical object.   
   \item \textbf{Franka-Emika cage}: 7-\textsc{dof} problem where the \textsc{franka-emika} robot starting outside a caged-box is reaching inside to pick up a cube.
   \item \textbf{Baxter  table}: 7-\textsc{dof} problem where the \textsc{baxter} robot starting with the arm under the table is reaching to pick a cylindrical object on a cluttered table. 
   \item \textbf{Fetch narrow bookshelf}: 8-\textsc{dof} problem where the \textsc{fetch} robot is starting from the home (stow) position and reaches for a cylindrical object deep in this narrow bookshelf. 
   \item \textbf{Baxter large bookshelf}: 14-\textsc{dof} problem where the \textsc{baxter} robot s starting outside the shelf reaches to pick two objects with both arms.
   \item \textbf{Shadow-Kuka kitchen}: 31-\textsc{dof} problem where a shadowhand mounted on a kuka arm starts inside the dish-washer and reaches inside the kitchen's shelf.
\end{enumerate}
As shown in Fig.~\ref{fig:experiments:manipulation}, RRT-Connect has the best overall success rate and is the only planner reaching $100\%$ in 5 out of 6 scenarios. Only KPIECE is able to reach $100\%$ quicker than RRTConnect in 1 out of 6 scenarios. In terms of cost, PRM* converges quickly to a low-cost solution in 4, while AIT* in 3 out of 6 scenarios. RRT* only finds a low-cost solution in 1 scenario.

\subsection{Narrow Passage Experiments}
\input{evaluations/Limitations/limitations_experiments}

This set of experiments includes six different types of narrow passages.
Narrow passages are difficult to solve for sampling-based motion planners, because the passages constitute bottlenecks, i.e. regions of the state space, which have near-zero measure, and near-zero probability to sample~states~\cite{Hsu2006probabilistic}. 
\begin{enumerate}
   \item \textbf{2d-Disk room}: A 2-\dof problem where a small disk moves through a slit in a wall.
   \item \textbf{Bugtrap}: A 6-\dof problem, where a cylindrical object (the bug) escapes a trap.
   \item \textbf{Double L-shape}: A 6-\dof problem, where an object consistent of two L-shapes has to traverse a square hole in a wall.
   \item \textbf{Rings on a helix}: A 6-\dof problem, where a ring has to move along a helix.
   \item \textbf{Franca-Emica box}: A  7-\dof problem, where a Franca-Emica robot has to move its endeffector inside of a box.
   \item \textbf{Twister snake}: A 10-\dof problem, where snake-like robot traverses a twisted pipe.
\end{enumerate}
In terms of success rate, the best planners are RRT-Connect and TRRT, which can solve 3 scenarios to almost $100\%$  success (see Fig.~\ref{fig:experiments:limitations}). While optimal planners can only find low-cost solutions in one scenario, is is noteworthy to mention that FMT can find low-cost solutions in five out of six scenarios. The twister scenario was not solved by any planner.

\subsection{Extension Experiments}
\input{evaluations/Extensions/extensions_experiments}

The extension experiments include contact-constraints, where robots are restricted to move along surfaces, and scenarios, where a dynamic model is used to actuate the robot. 
\begin{enumerate}
   \item \textbf{Implicit chain}: A 5-\dof problem where the end-effector of an articulated chain is bounded to the surface of a sphere~\cite{kingston2019}.
    \item \textbf{Constrained sphere}: A 3-\dof point robot which is bounded to the surface of a sphere with obstacles~\cite{kingston2019}.
    \item \textbf{Maze on Torus}: A 3-\dof point robot which is bounded to the surface of a torus plus a maze-like obstacle~\cite{kingston2019}.
    \item \textbf{Dubin's car}: A 3-\dof simple car model which can drive forwards at a constant speed (using a steering function).
    \item \textbf{Dynamic car}: A 3-\dof dynamic model of a car with a 2-\dof action space, which can accelerate and steer (using a propagator function).
    \item \textbf{Single-thruster UAV}: A 9-\dof dynamic model of an unmanned aerial vehicle (UAV), which has a single actuated thruster which can be pivoted.
\end{enumerate}
Fig.~\ref{fig:experiments:extensions} shows the result. Using the constraint-based framework~\cite{kingston2019}, EST, BIT*, and RRT-Connect can solve two scenarios to $100 \%$ , while BIT* finds low-cost solutions in three. In the dynamic cases, only Kinodynamic-RRT is able to find solutions, while the optimal planner SST* cannot solve any of the scenarios.

%% file: evaluations/Classical/classical_experiments.tex
\begin{figure*}
    \centering
\def\width{0.16\linewidth}
\begin{subfigure}[t]{\width}
\addSubCaption{Barriers}
\centering
\includegraphics[width=\linewidth]{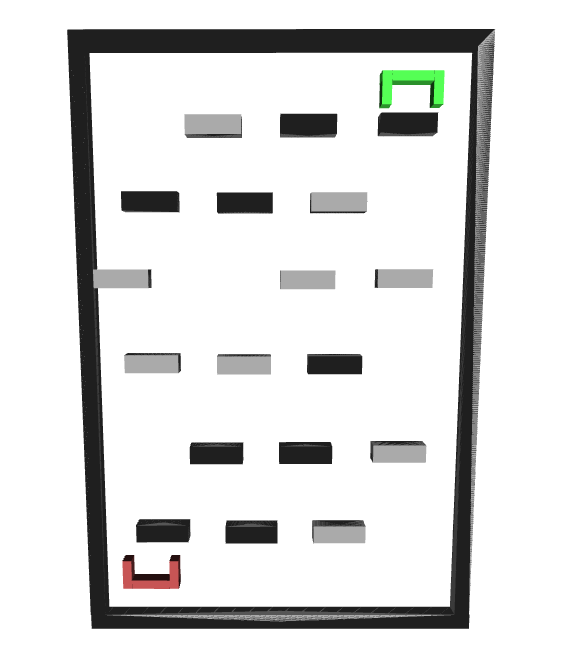}
\end{subfigure}
\begin{subfigure}[t]{\width}
\addSubCaption{Maze}
\centering
\includegraphics[width=\linewidth]{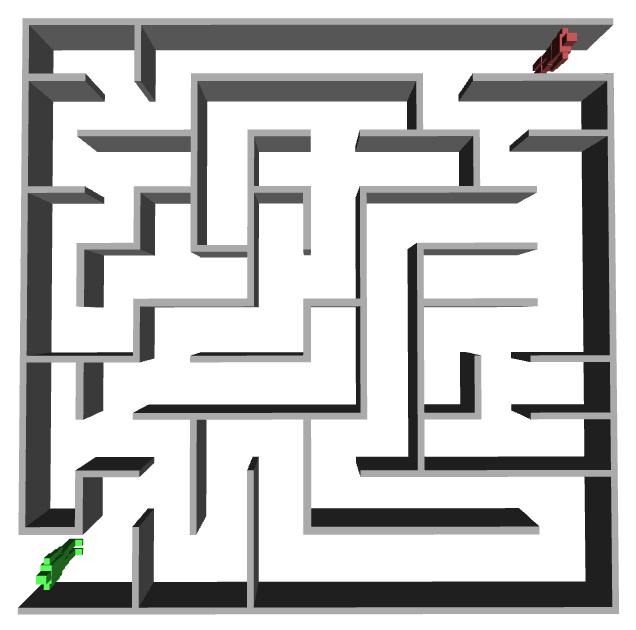}
\end{subfigure}
\begin{subfigure}[t]{\width}
\addSubCaption{Polygons}
\centering
\includegraphics[width=\linewidth]{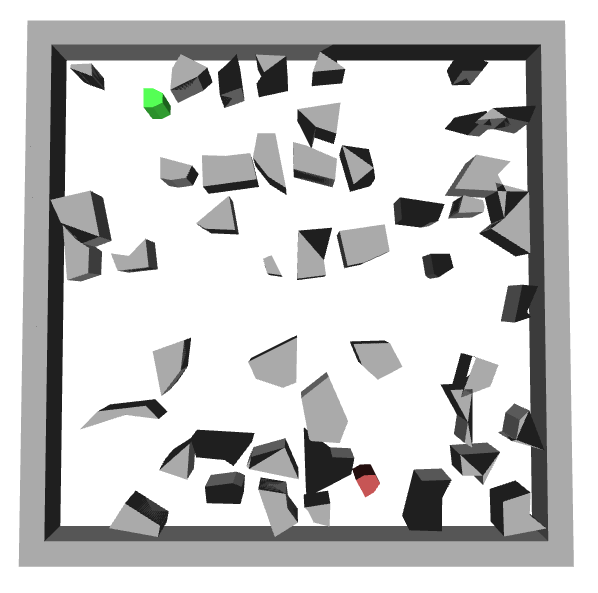}
\end{subfigure}
\begin{subfigure}[t]{\width}
\addSubCaption{Apartment}
\centering
\includegraphics[width=\linewidth]{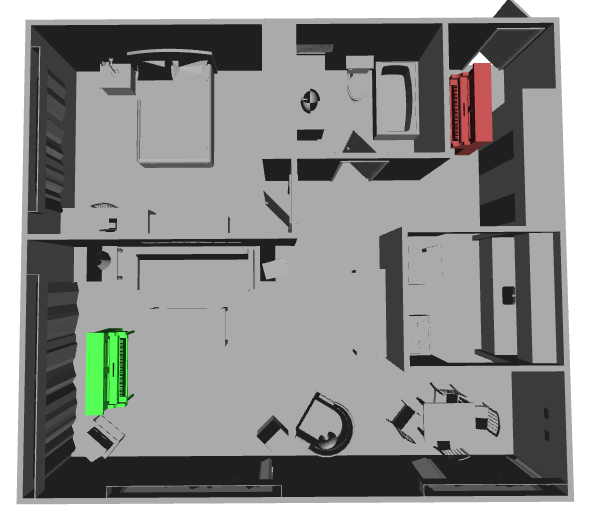}
\end{subfigure}
\begin{subfigure}[t]{\width}
\addSubCaption{Cubicles}
\centering
\includegraphics[width=\linewidth]{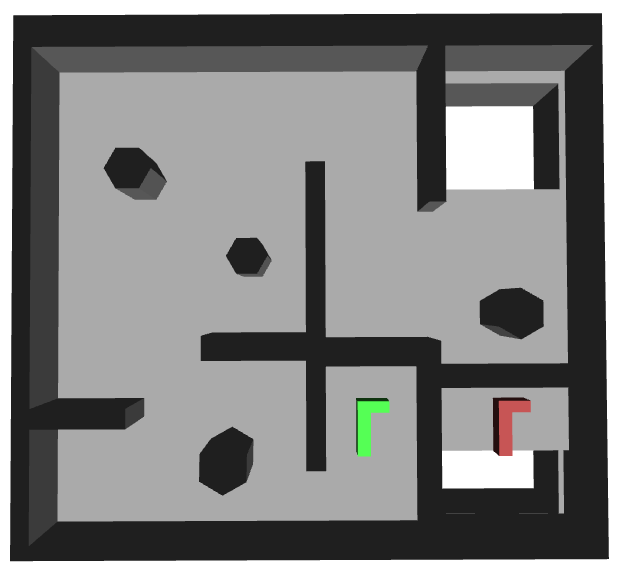}
\end{subfigure}
\begin{subfigure}[t]{\width}
\addSubCaption{Home}
\centering
\includegraphics[width=\linewidth]{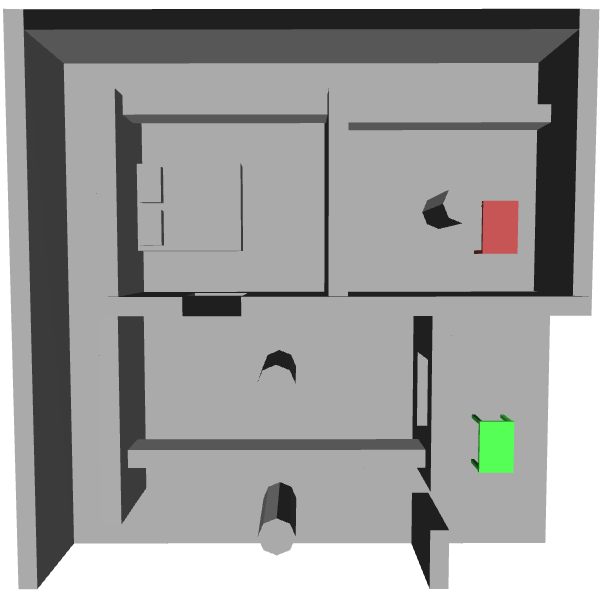}
\end{subfigure}
\def\width{0.32\linewidth}
\begin{subfigure}[t]{\width}
\centering
\includegraphics[width=\linewidth]{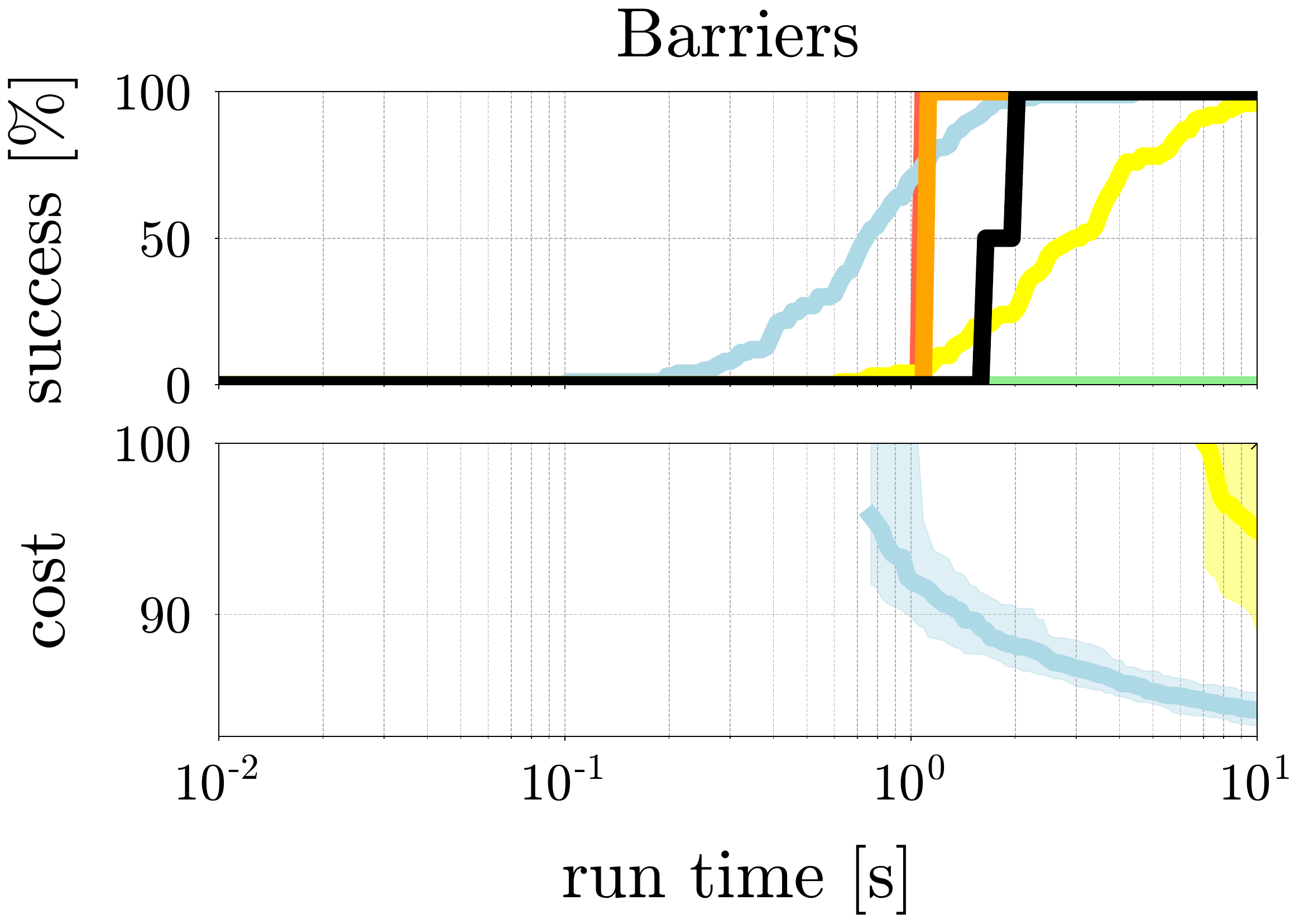}
\end{subfigure}
\begin{subfigure}[t]{\width}
\centering
\includegraphics[width=\linewidth]{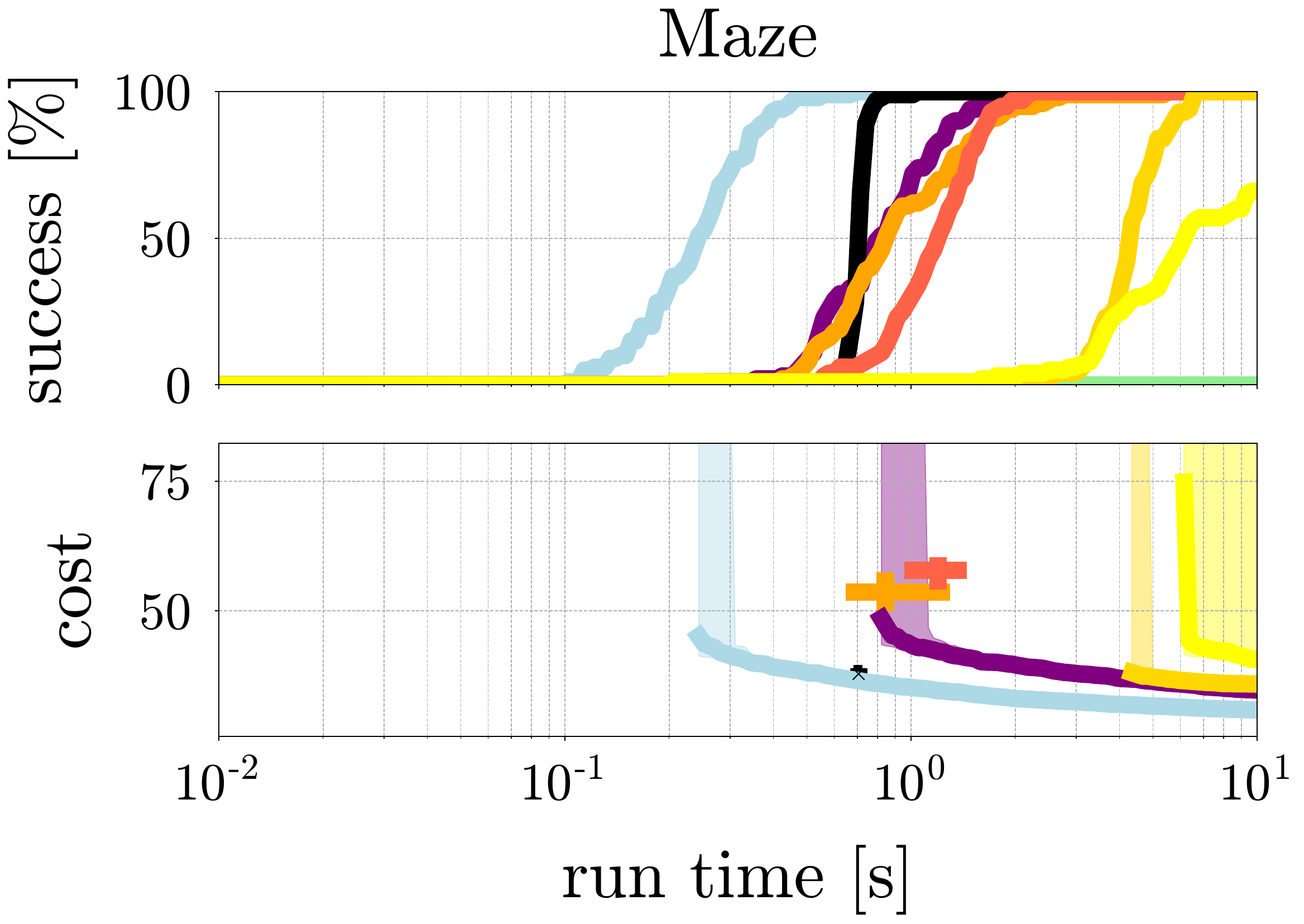}
\end{subfigure}
\begin{subfigure}[t]{\width}
\centering
\includegraphics[width=\linewidth]{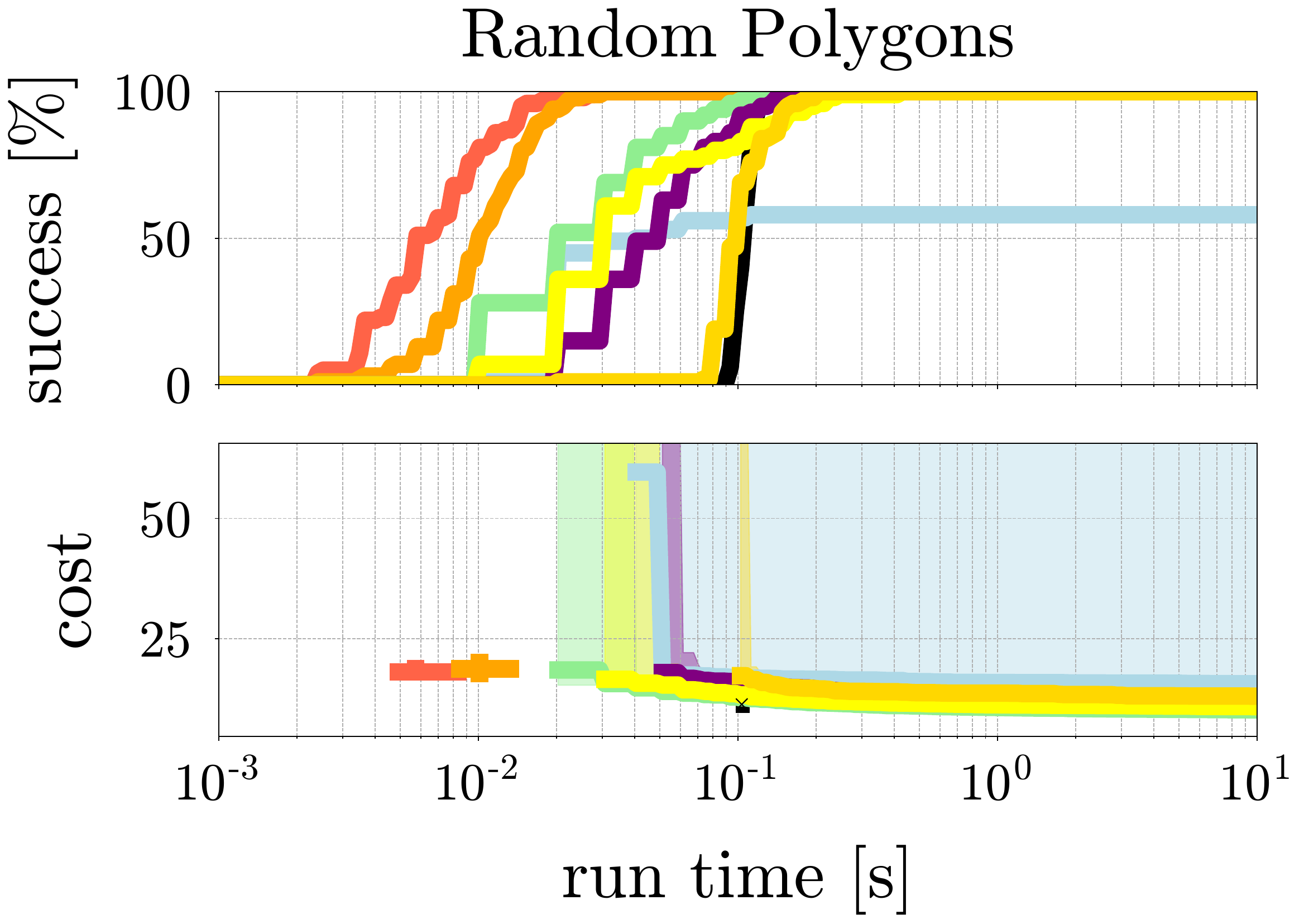}
\end{subfigure}
\begin{subfigure}[t]{\width}
\centering
\includegraphics[width=\linewidth]{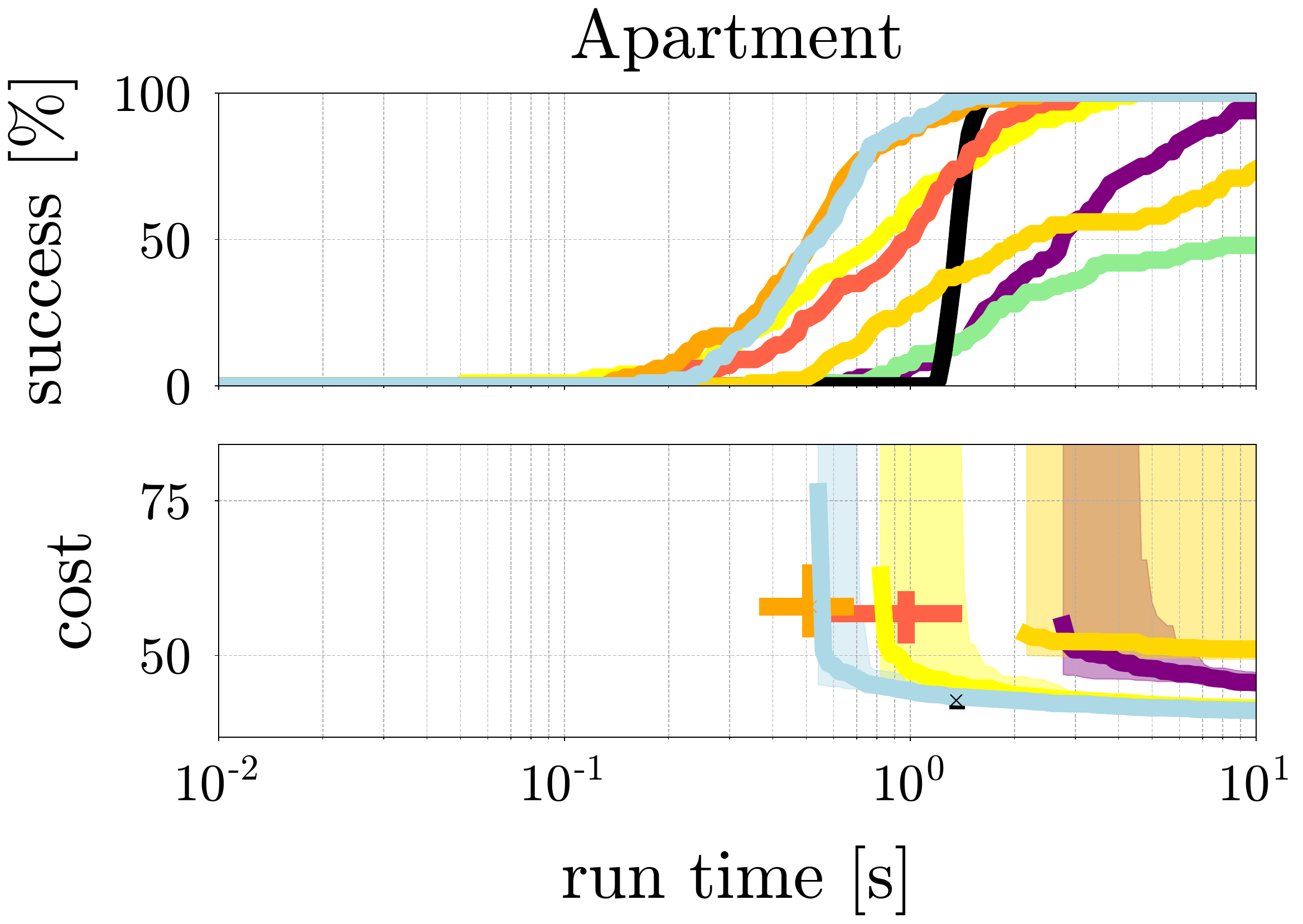}
\end{subfigure}
\begin{subfigure}[t]{\width}
\centering
\includegraphics[width=\linewidth]{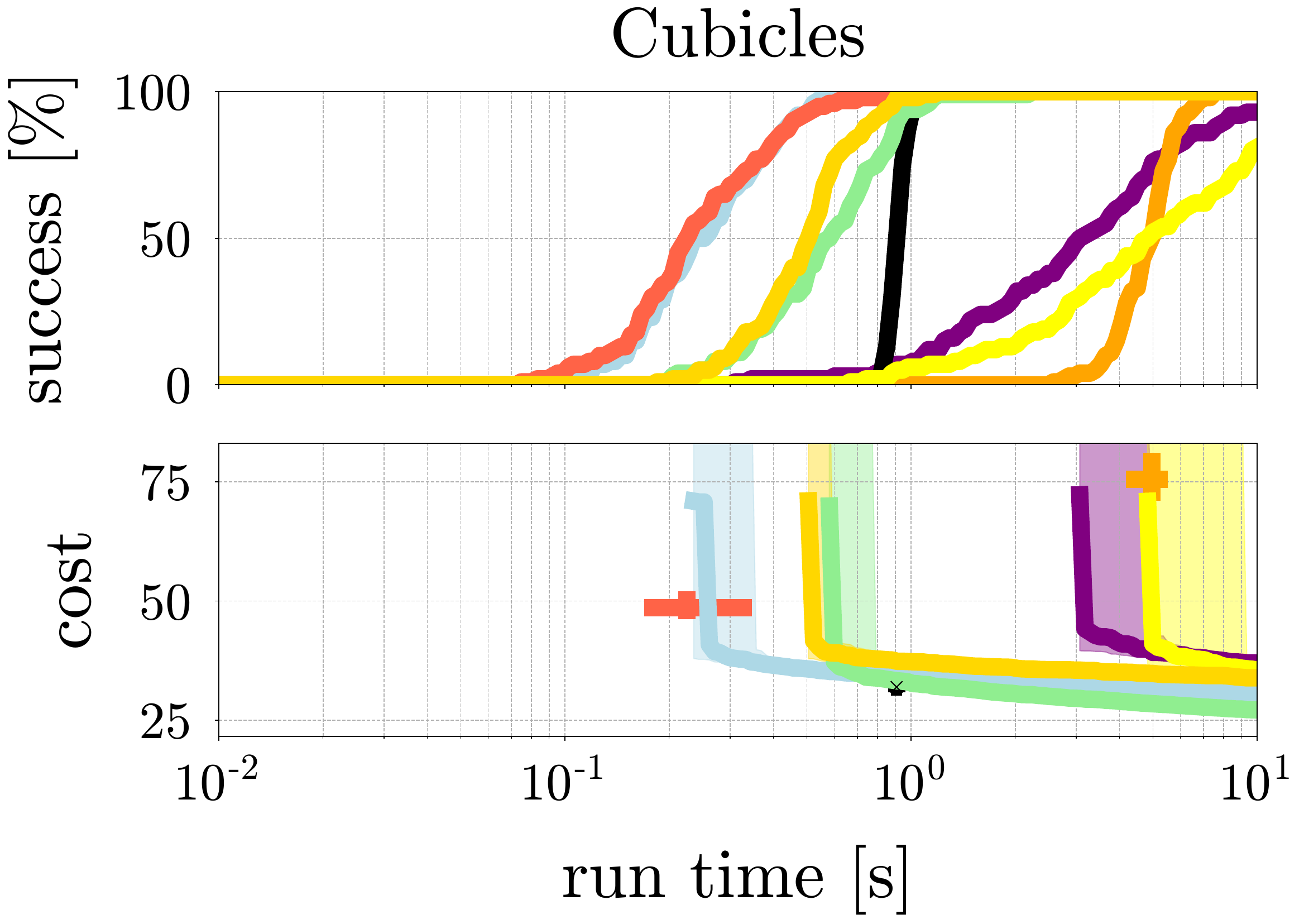}
\end{subfigure}
\begin{subfigure}[t]{\width}
\centering
\includegraphics[width=\linewidth]{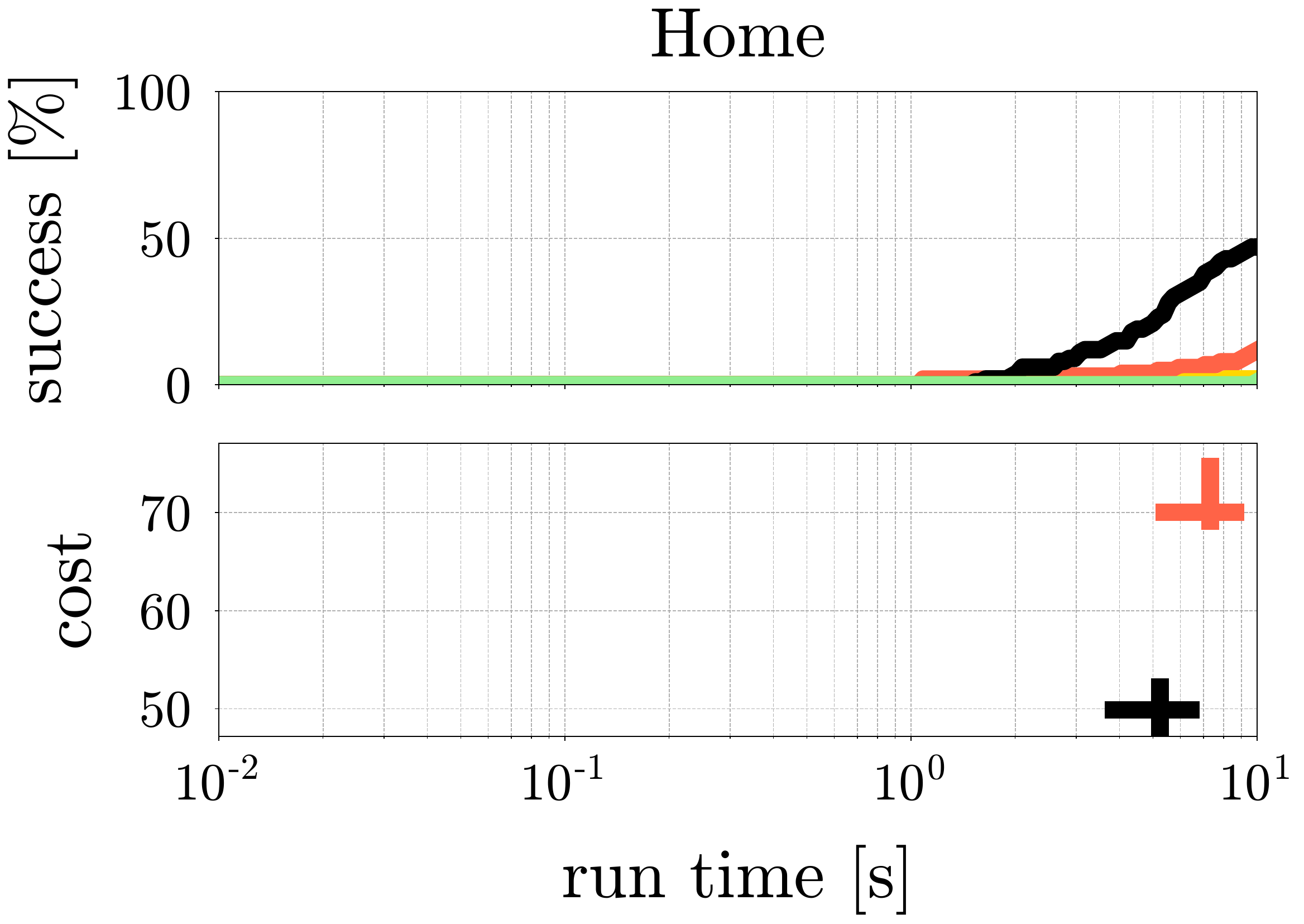}
\end{subfigure}
\caption{Classical Experiments with planners \sqbox{crrtconnect} RRT-Connect, \sqbox{cprm} PRM, \sqbox{crrtstar} RRT*, \sqbox{cfmt} FMT, \sqbox{cest} EST, \sqbox{clbtrrt} LBTRRT, \sqbox{cait} AIT*, and \sqbox{cbitstar} BIT*.
\label{fig:experiments:classical}}
\vspace{-1em}
\end{figure*}

%% file: evaluations/Manipulation/manipulation_experiments.tex
\begin{figure*}
    \centering
\def\width{0.16\linewidth}
\begin{subfigure}[t]{\width}
\addSubCaption{Baxter Shelf}
\centering
\includegraphics[width=\linewidth]{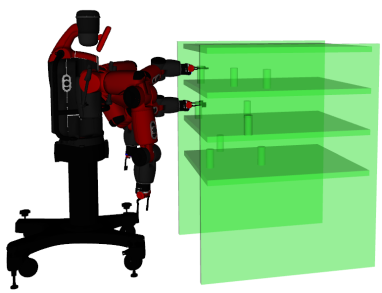}
\end{subfigure}
\begin{subfigure}[t]{\width}
\addSubCaption{Fetch Thin}
\centering
\includegraphics[width=\linewidth]{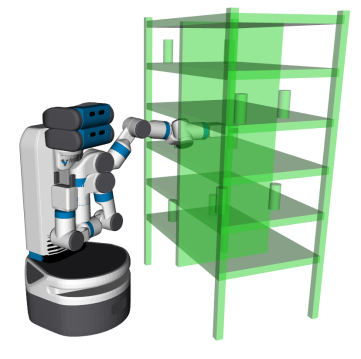}
\end{subfigure}
\begin{subfigure}[t]{\width}
\addSubCaption{UR5 Shelf}
\centering
\includegraphics[width=\linewidth]{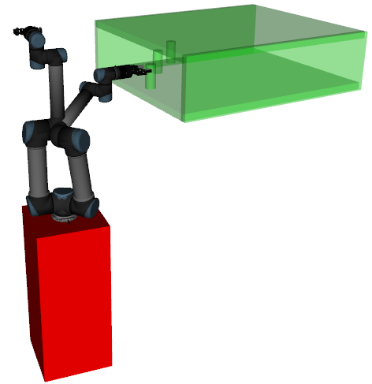}
\end{subfigure}
\begin{subfigure}[t]{\width}
\addSubCaption{Shadow Kitchen}
\centering
\includegraphics[width=\linewidth]{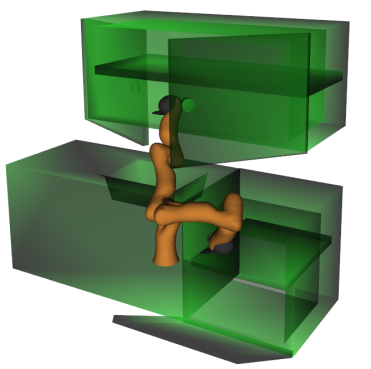}
\end{subfigure}
\begin{subfigure}[t]{\width}
\addSubCaption{Panda Cage}
\centering
\includegraphics[width=\linewidth]{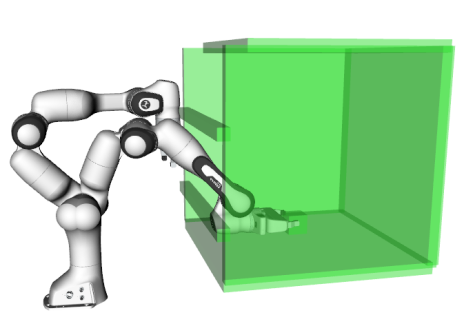}
\end{subfigure}
\begin{subfigure}[t]{\width}
\addSubCaption{Baxter Table}
\centering
\includegraphics[width=\linewidth]{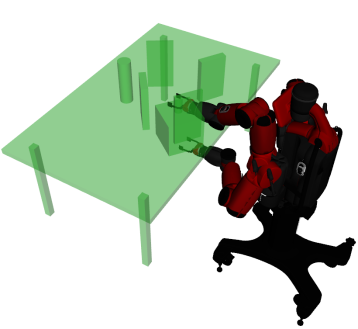}
\end{subfigure}

\def\width{0.32\linewidth}
\begin{subfigure}[t]{\width}
\centering
\includegraphics[width=\linewidth]{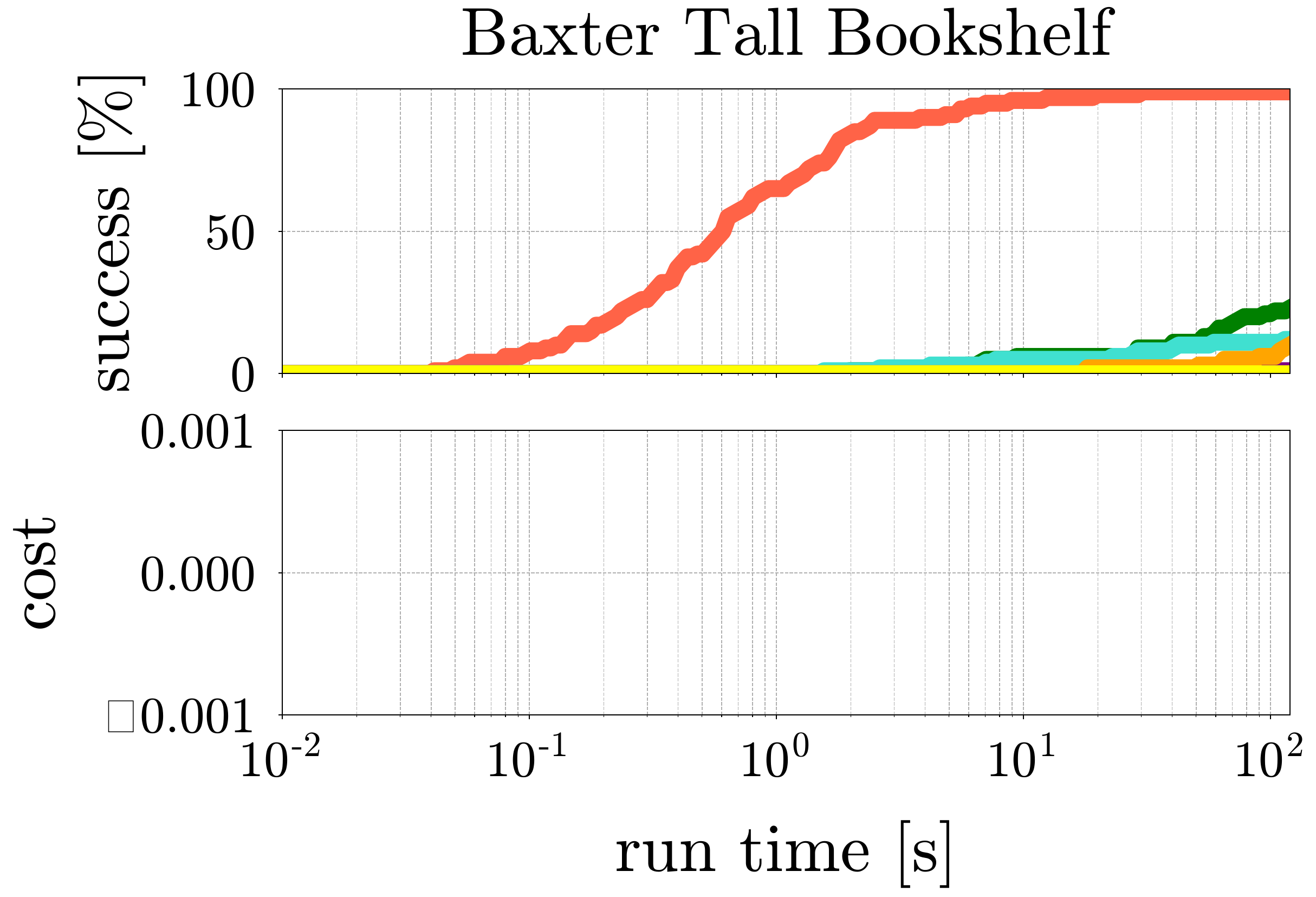}
\end{subfigure}
\begin{subfigure}[t]{\width}
\centering
\includegraphics[width=\linewidth]{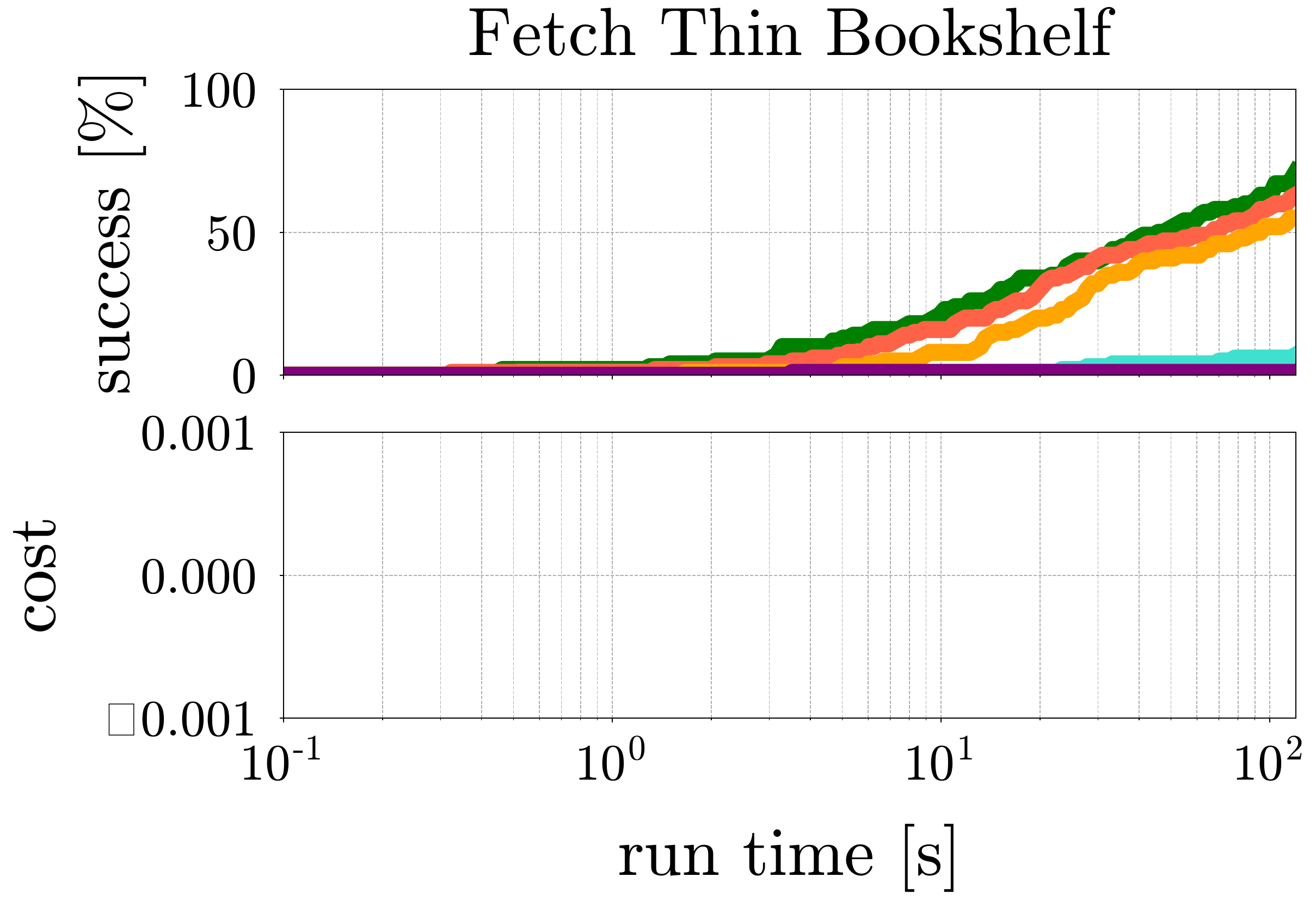}
\end{subfigure}
\begin{subfigure}[t]{\width}
\centering
\includegraphics[width=\linewidth]{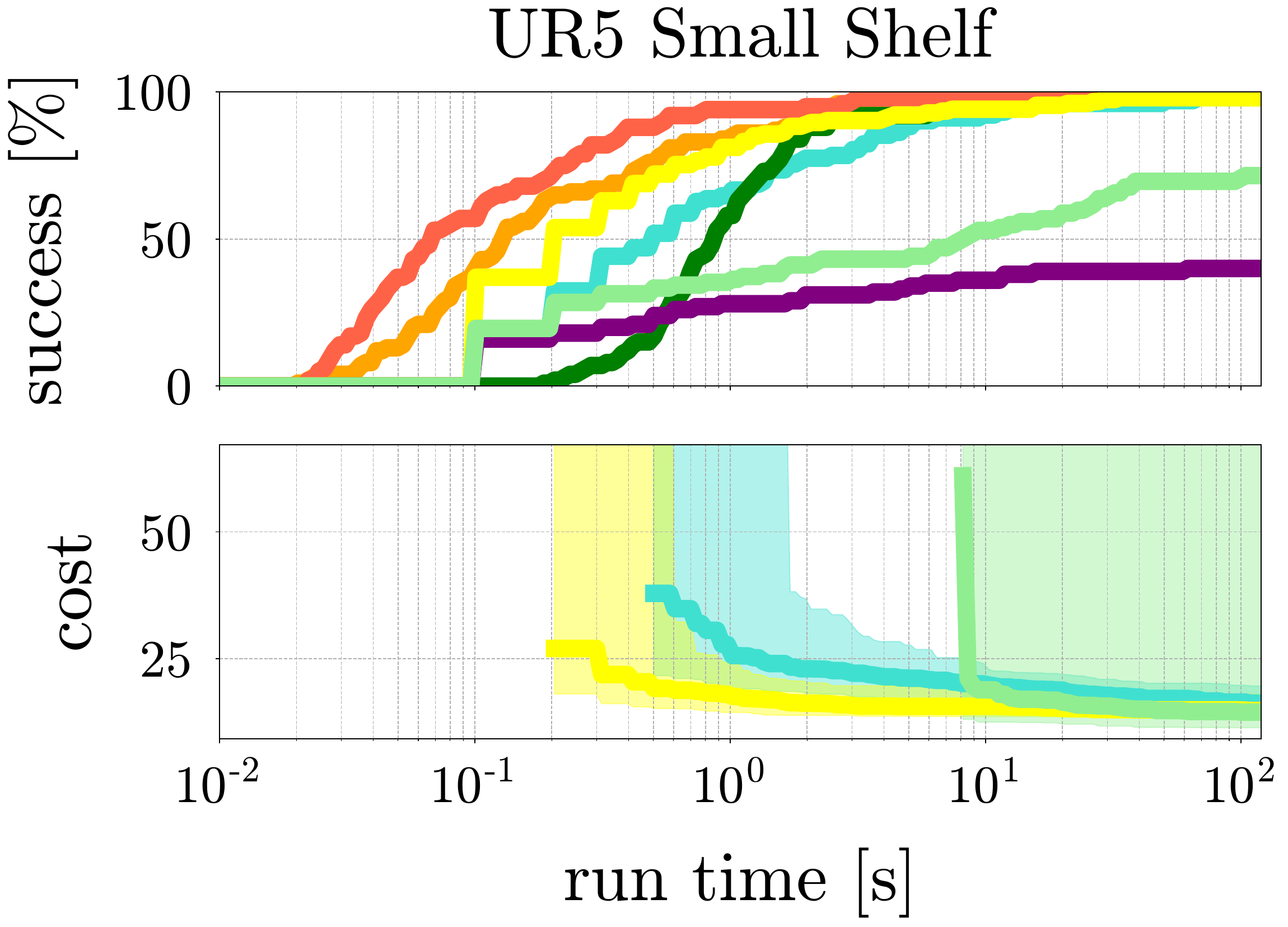}
\end{subfigure}
\begin{subfigure}[t]{\width}
\centering
\includegraphics[width=\linewidth]{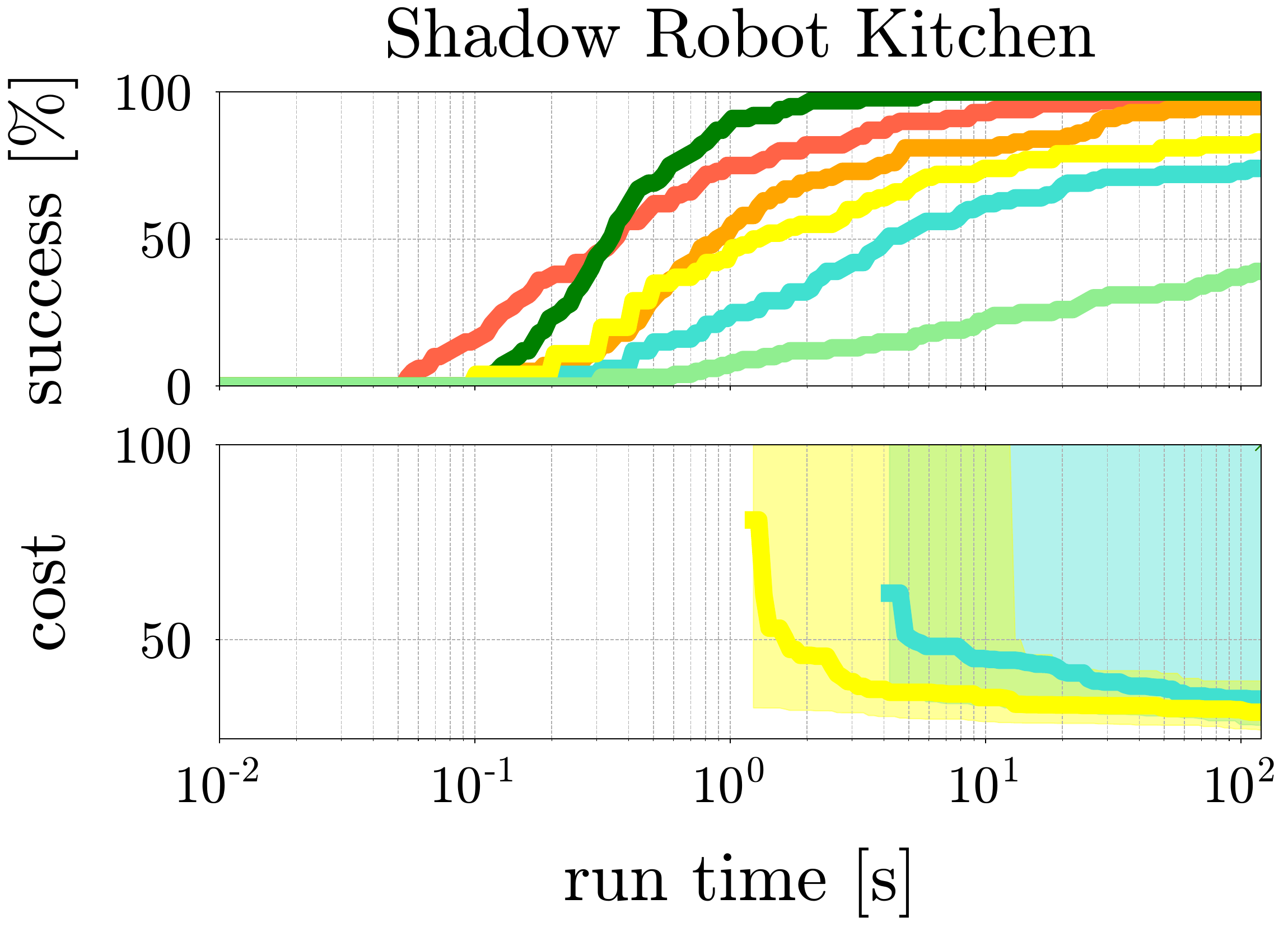}
\end{subfigure}
\begin{subfigure}[t]{\width}
\centering
\includegraphics[width=\linewidth]{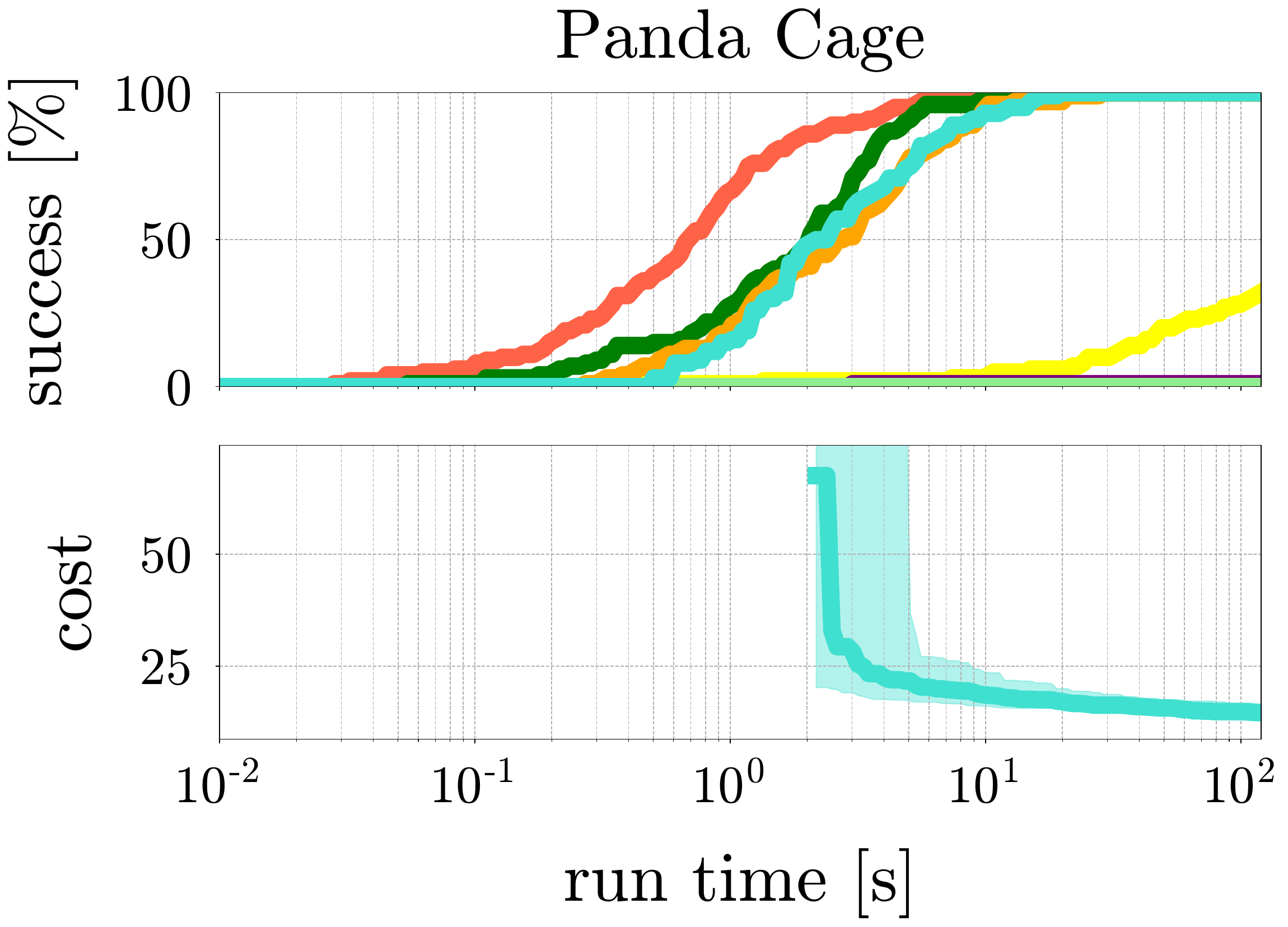}
\end{subfigure}
\begin{subfigure}[t]{\width}
\centering
\includegraphics[width=\linewidth]{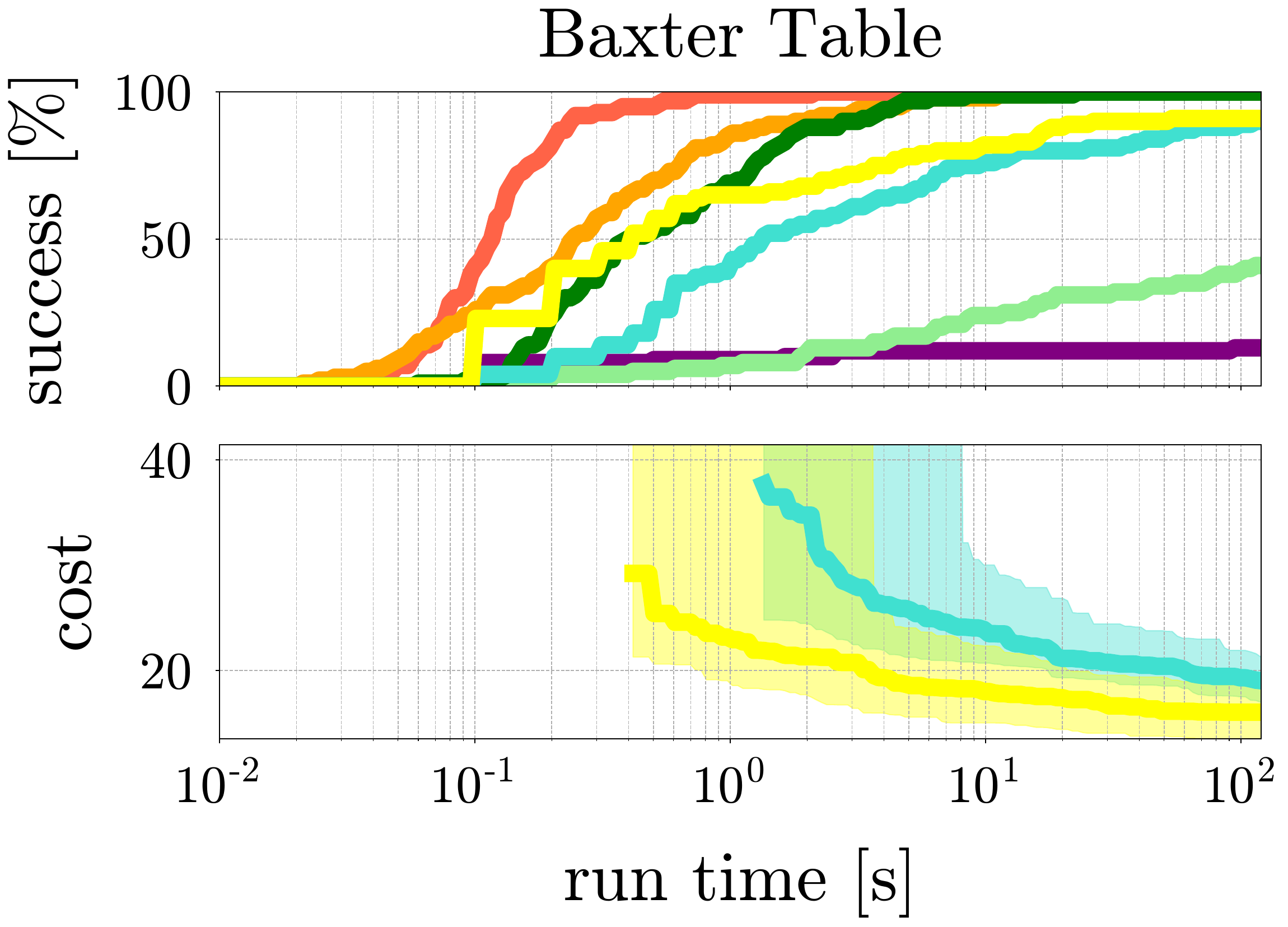}
\end{subfigure}
\caption{Manipulation experiments using planners \sqbox{crrtconnect}~RRTConnect, \sqbox{cprm}~PRM, \sqbox{cprmstar}~PRM*, \sqbox{crrtstar}~RRT*, \sqbox{cest}~EST, \sqbox{ckpiece}~KPIECE, and \sqbox{cait}~AIT*\label{fig:experiments:manipulation}.}
\vspace{-2em}
\end{figure*}

%% file: evaluations/Limitations/limitations_experiments.tex
\begin{figure*}[h!]
    \centering
\def\width{0.16\linewidth}
\begin{subfigure}[t]{\width}
\addSubCaption{Disk}
\centering
\includegraphics[width=\linewidth]{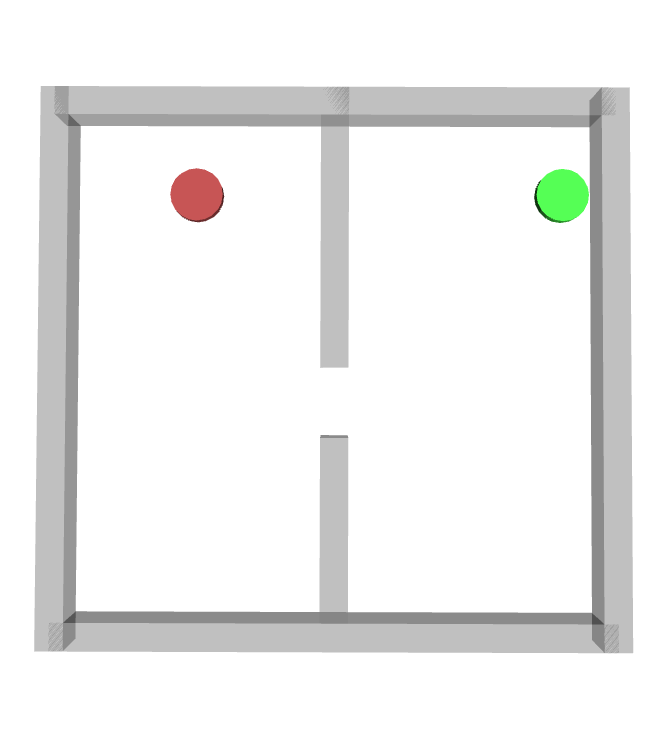}
\end{subfigure}
\begin{subfigure}[t]{\width}
\addSubCaption{Bugtrap}
\centering
\includegraphics[width=\linewidth]{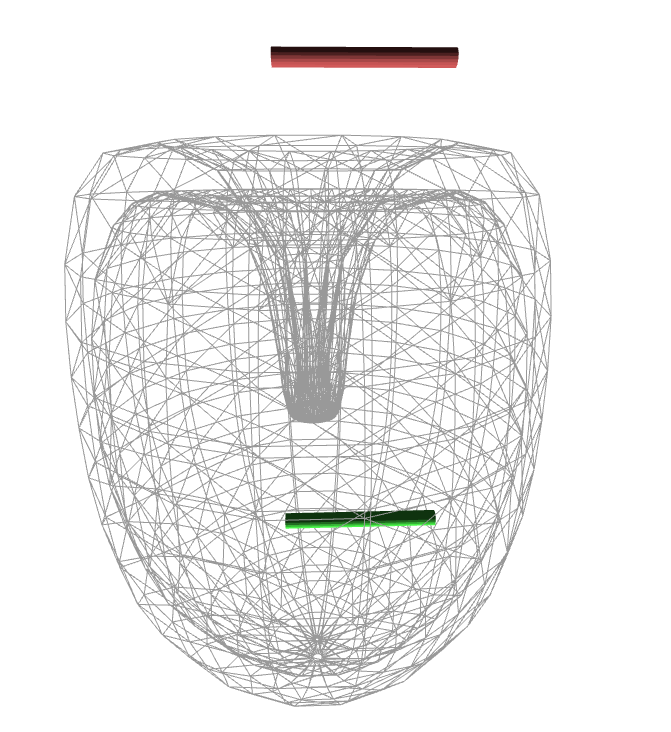}
\end{subfigure}
\begin{subfigure}[t]{\width}
\addSubCaption{Double L}
\centering
\includegraphics[width=\linewidth]{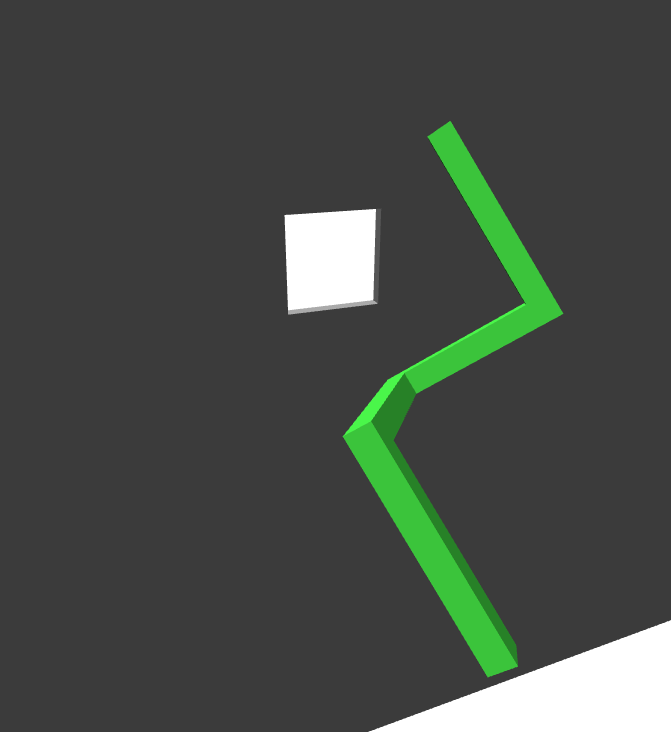}
\end{subfigure}
\begin{subfigure}[t]{\width}
\addSubCaption{Rings}
\centering
\includegraphics[width=\linewidth]{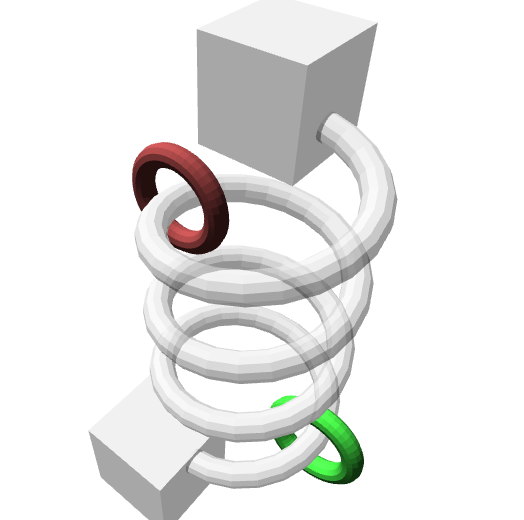}
\end{subfigure}
\begin{subfigure}[t]{\width}
\addSubCaption{Panda}
\centering
\includegraphics[width=\linewidth]{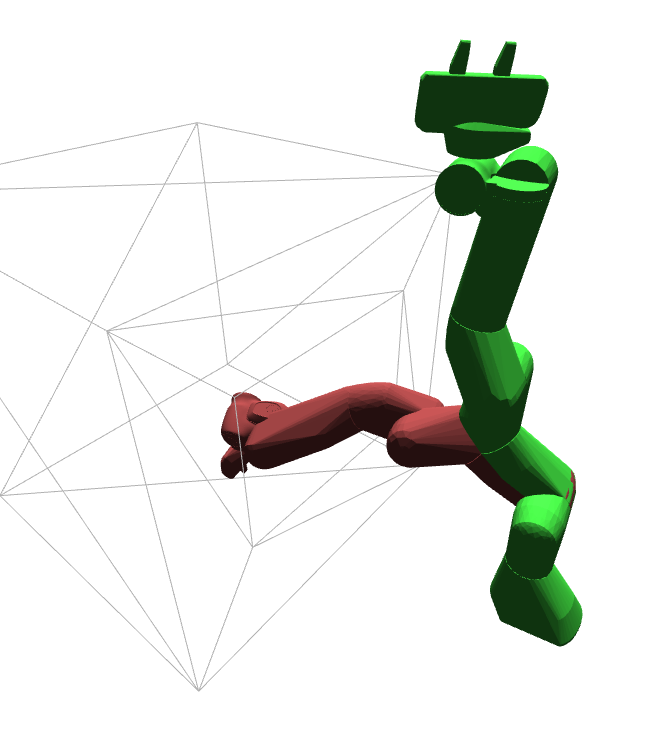}
\end{subfigure}
\begin{subfigure}[t]{\width}
\addSubCaption{Twister}
\centering
\includegraphics[width=\linewidth]{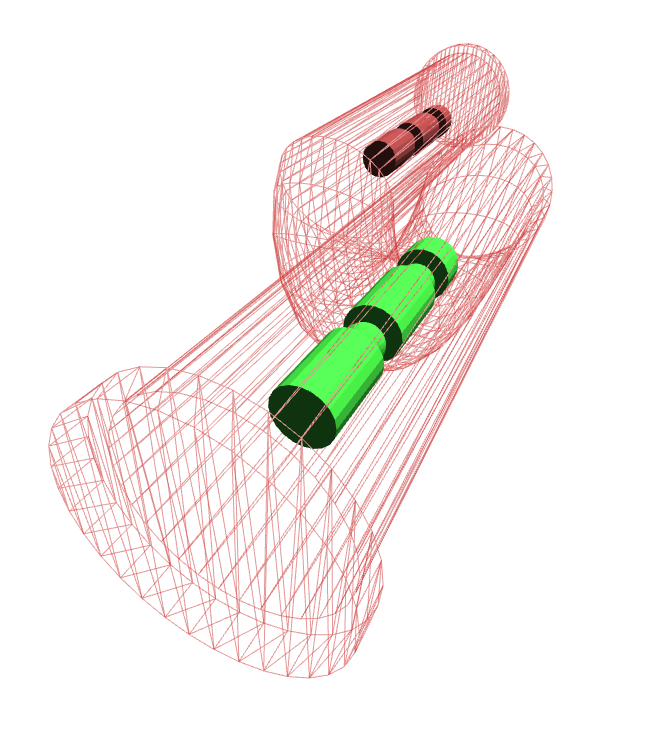}
\end{subfigure}
\def\width{0.32\linewidth}
\begin{subfigure}[t]{\width}
\centering
\includegraphics[width=\linewidth]{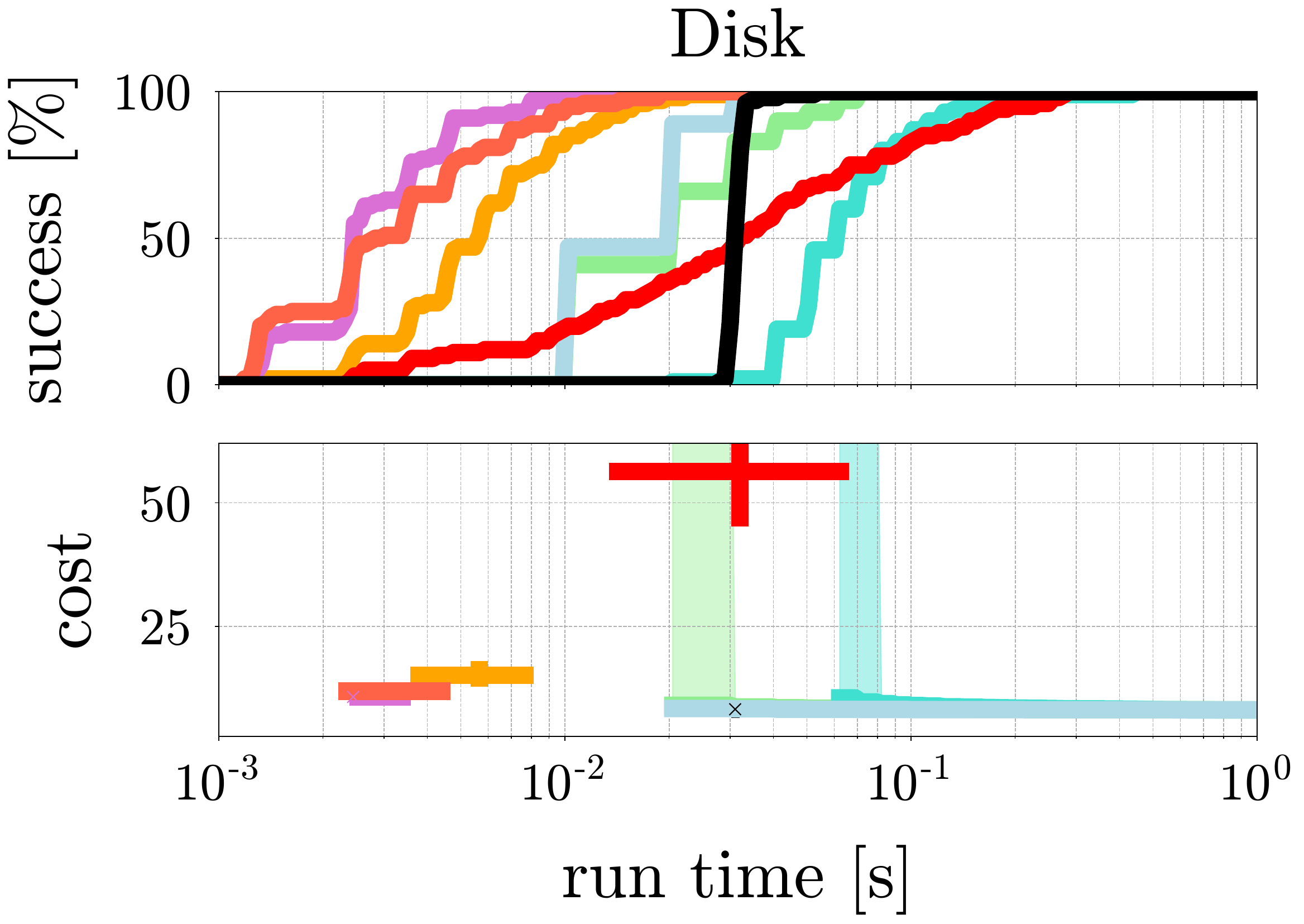}
\end{subfigure}
\begin{subfigure}[t]{\width}
\centering
\includegraphics[width=\linewidth]{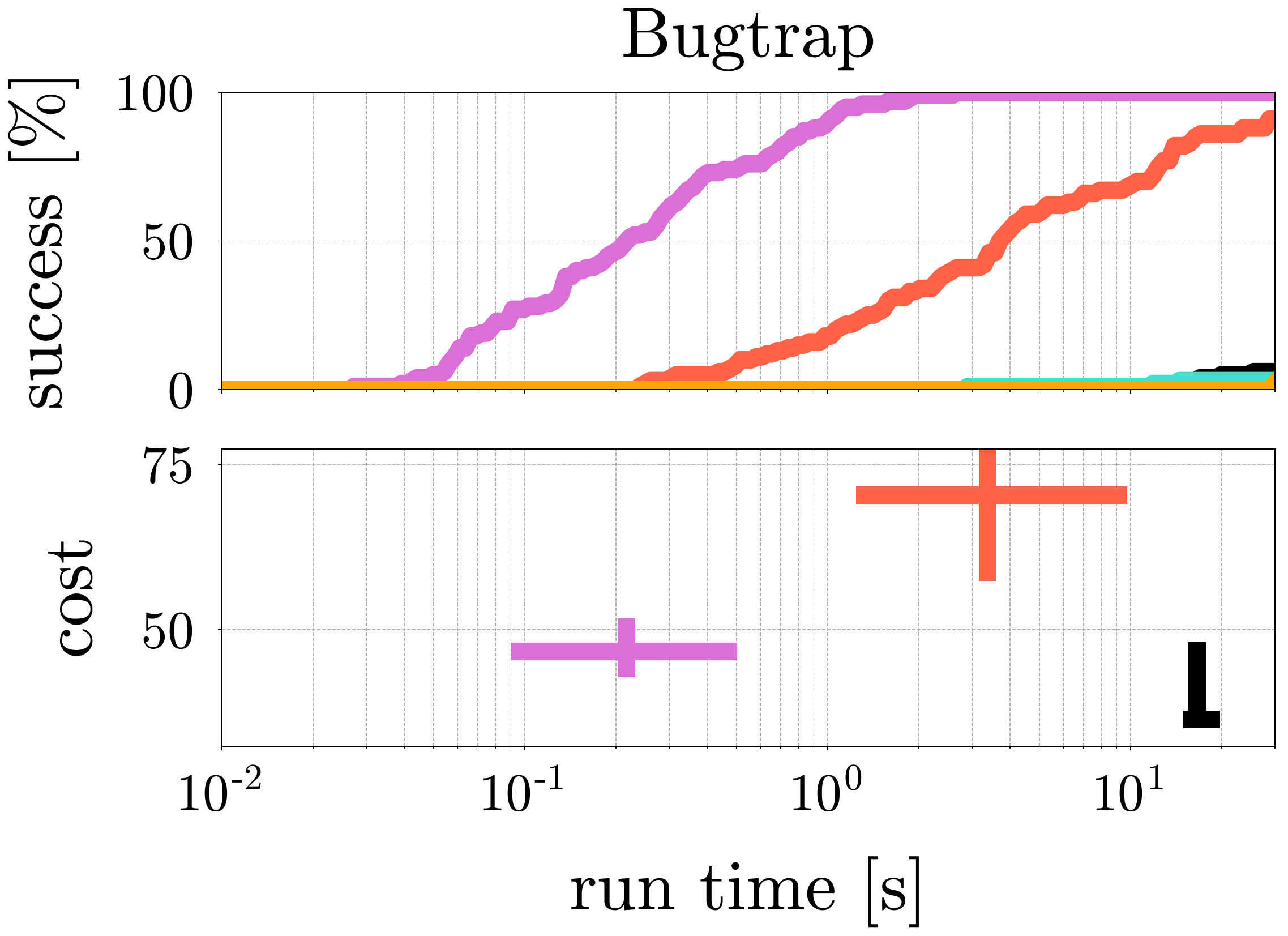}
\end{subfigure}
\begin{subfigure}[t]{\width}
\centering
\includegraphics[width=\linewidth]{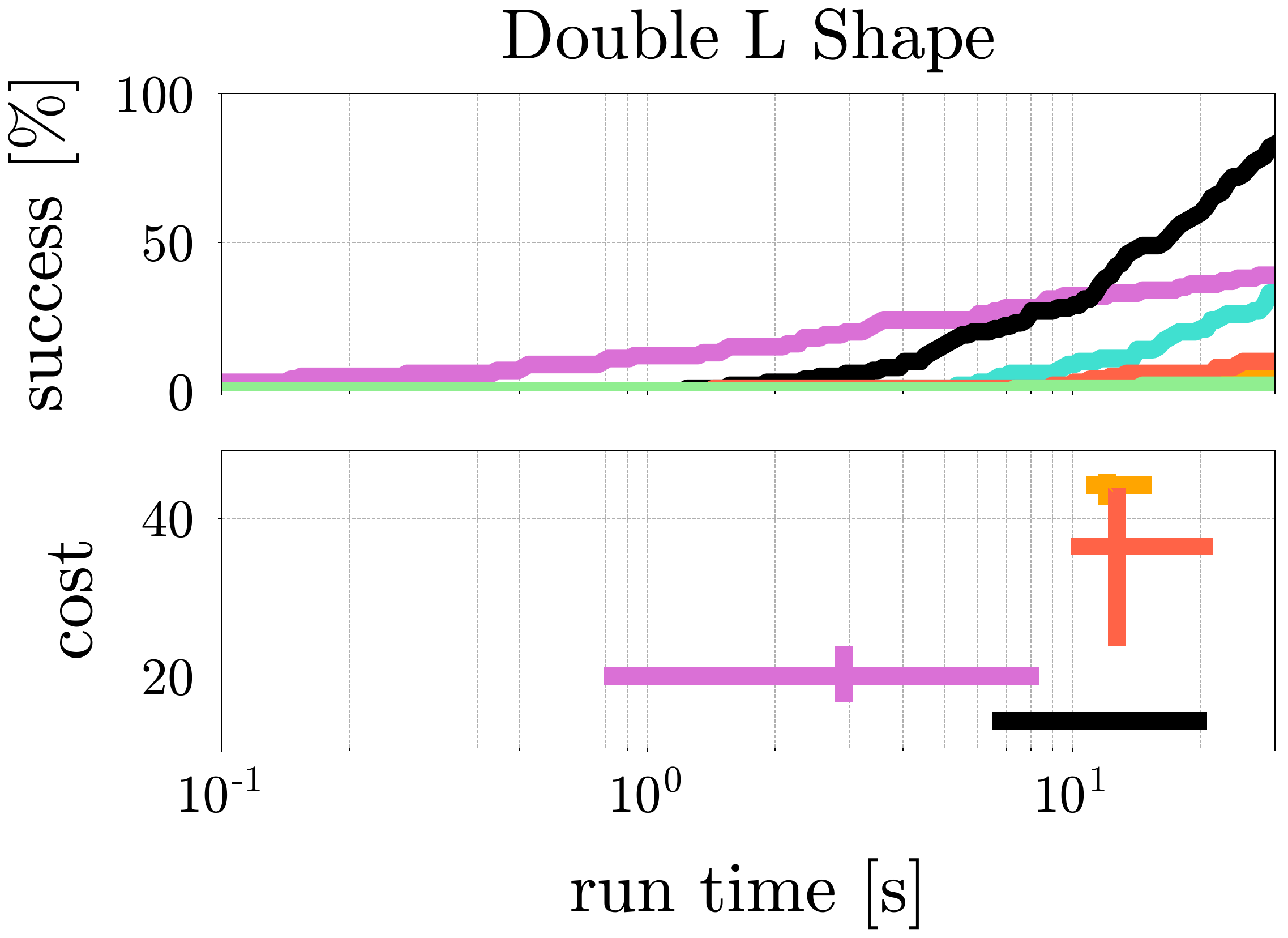}
\end{subfigure}
\begin{subfigure}[t]{\width}
\centering
\includegraphics[width=\linewidth]{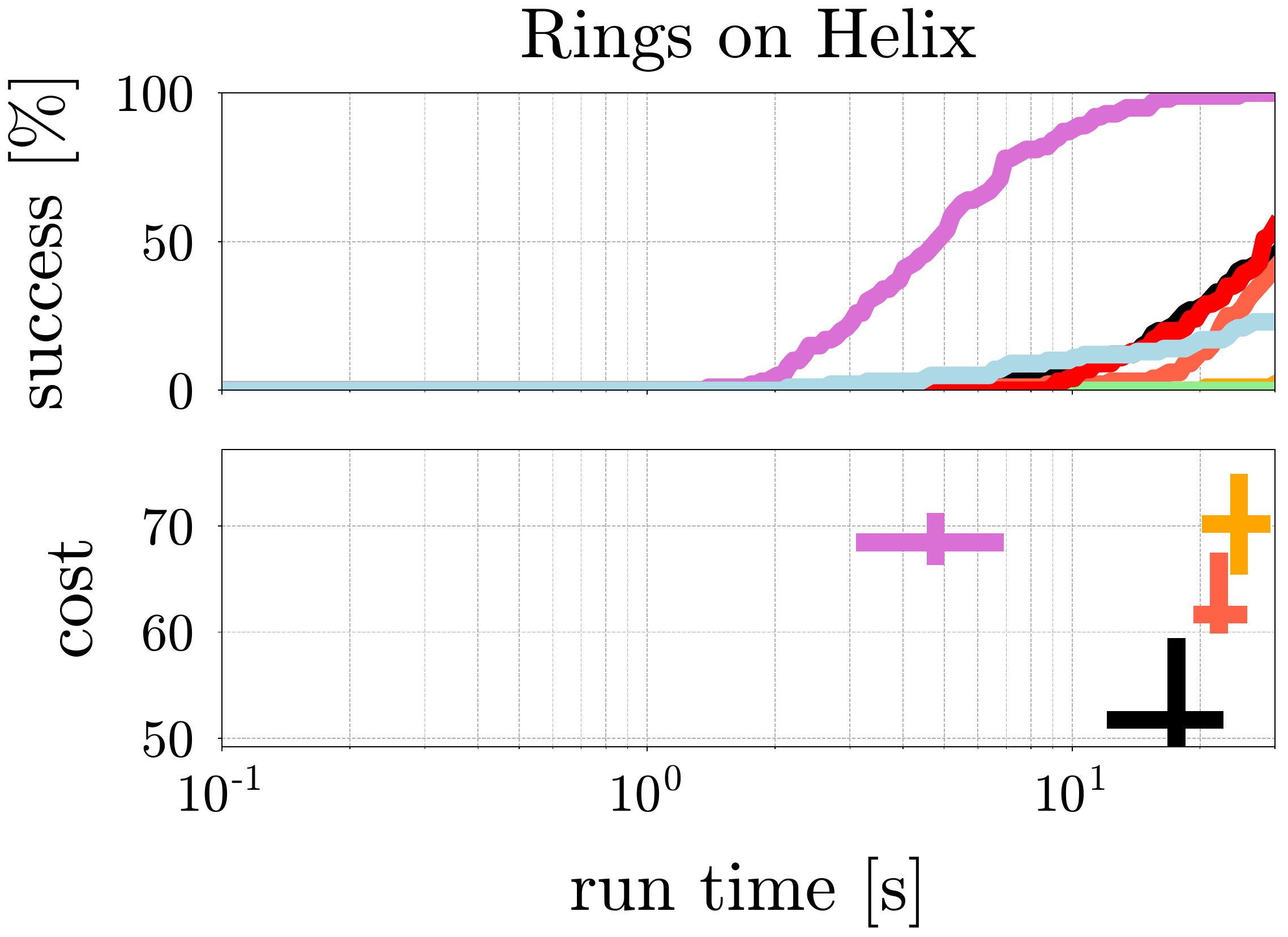}
\end{subfigure}
\begin{subfigure}[t]{\width}
\centering
\includegraphics[width=\linewidth]{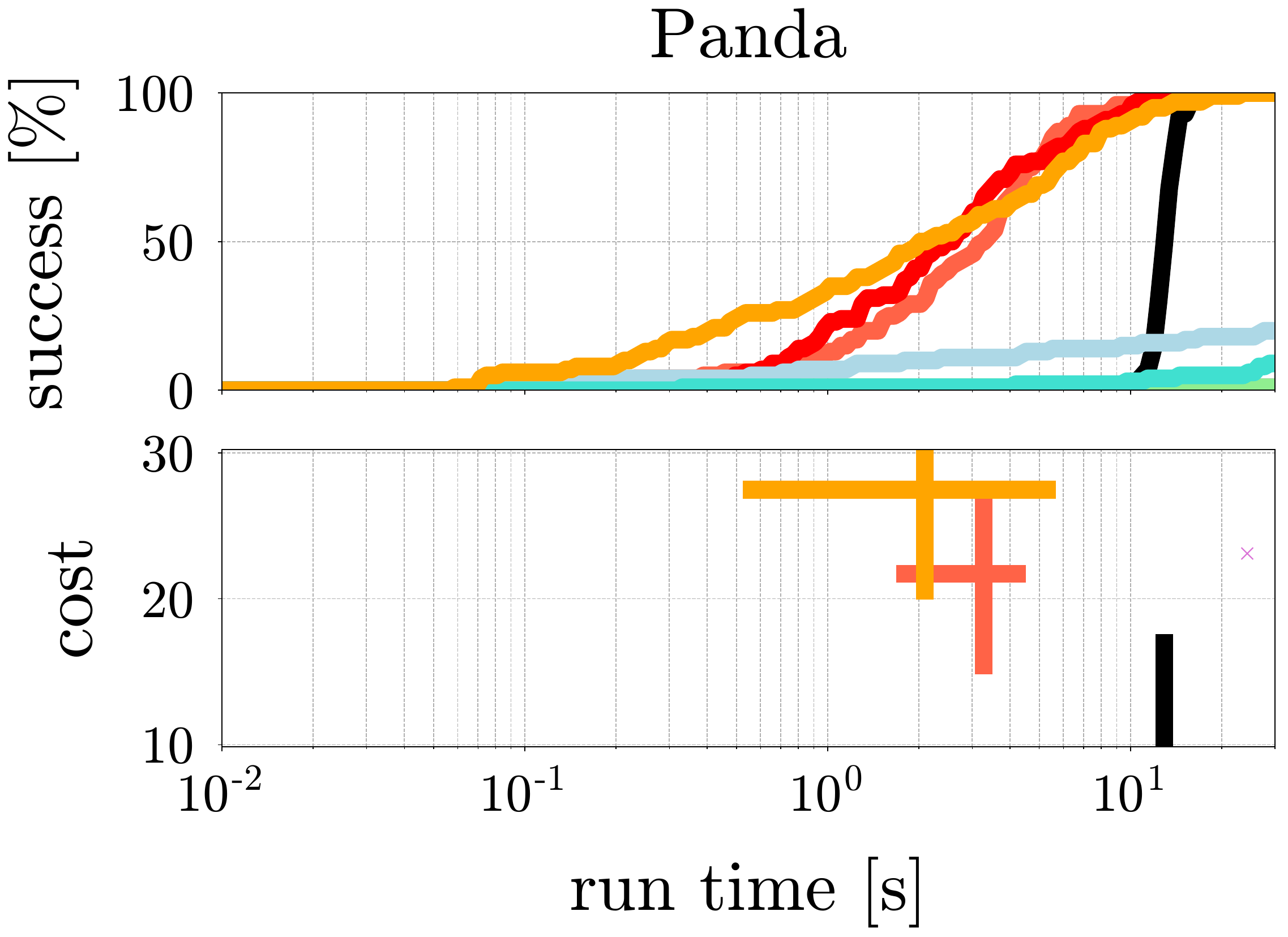}
\end{subfigure}
\begin{subfigure}[t]{\width}
\centering
\includegraphics[width=\linewidth]{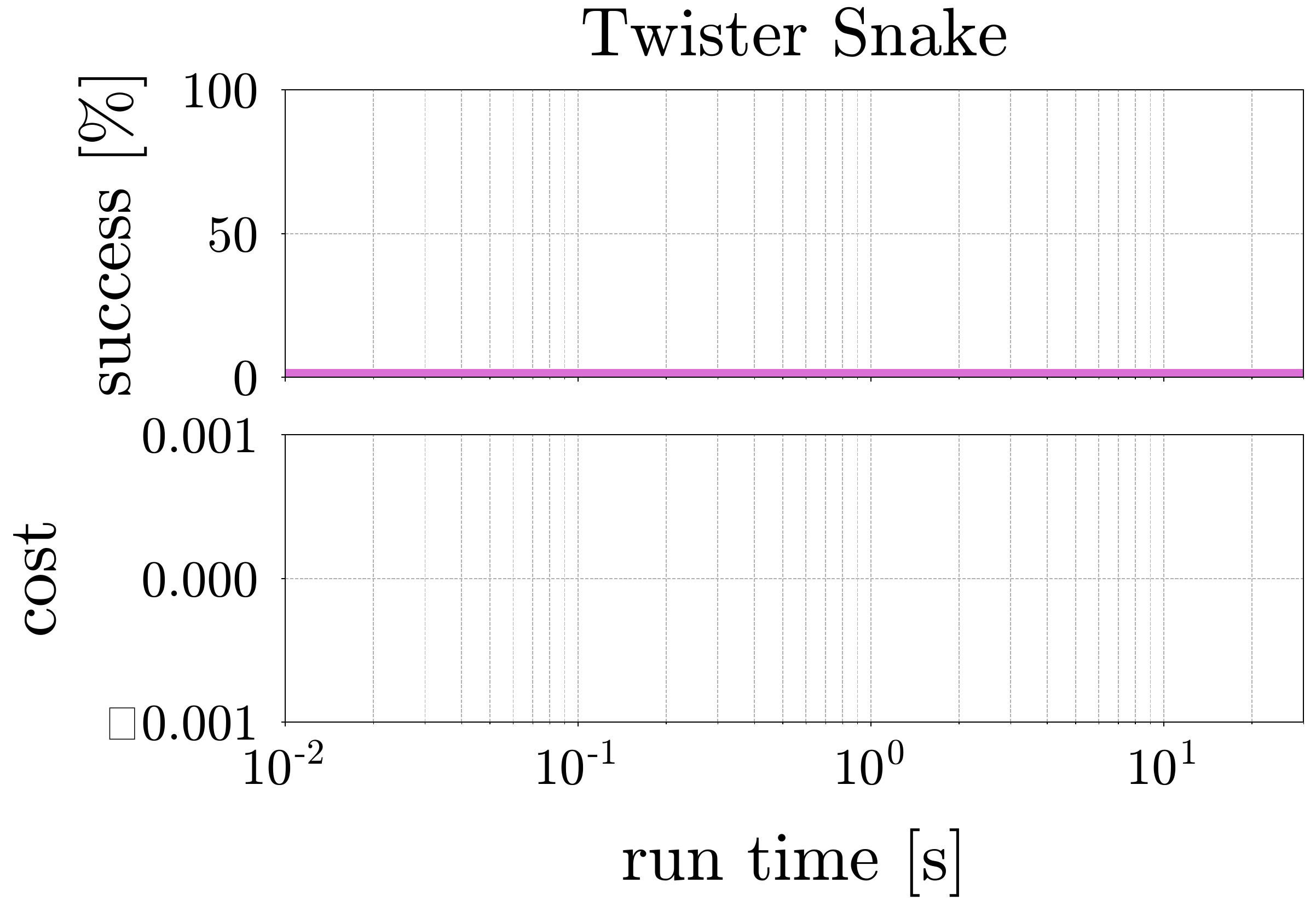}
\end{subfigure}
\caption{Limitations Experiments. Using planners \sqbox{crrtconnect}~RRT-Connect, \sqbox{cprmstar}~PRM*, \sqbox{crrtstar}~RRT*, \sqbox{ckpiece}~KPIECE, \sqbox{cfmt}~FMT, \sqbox{cest}~EST, \sqbox{ctrrt}~TRRT and \sqbox{cbitstar}~BIT*. \label{fig:experiments:limitations}}
\vspace{-1em}
\end{figure*}

%% file: evaluations/Extensions/extensions_experiments.tex
\begin{figure*}[h!]
    \centering
\def\width{0.16\linewidth}
\begin{subfigure}[t]{\width}
\addSubCaption{Implicit}
\centering
\includegraphics[width=\linewidth]{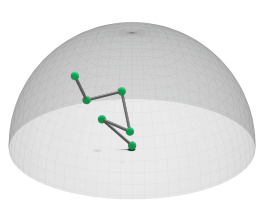}
\end{subfigure}
\begin{subfigure}[t]{\width}
\addSubCaption{Sphere}
\centering
\includegraphics[width=\linewidth]{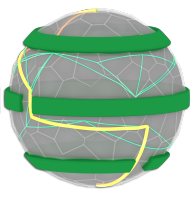}
\end{subfigure}
\begin{subfigure}[t]{\width}
\addSubCaption{Torus}
\centering
\includegraphics[width=\linewidth]{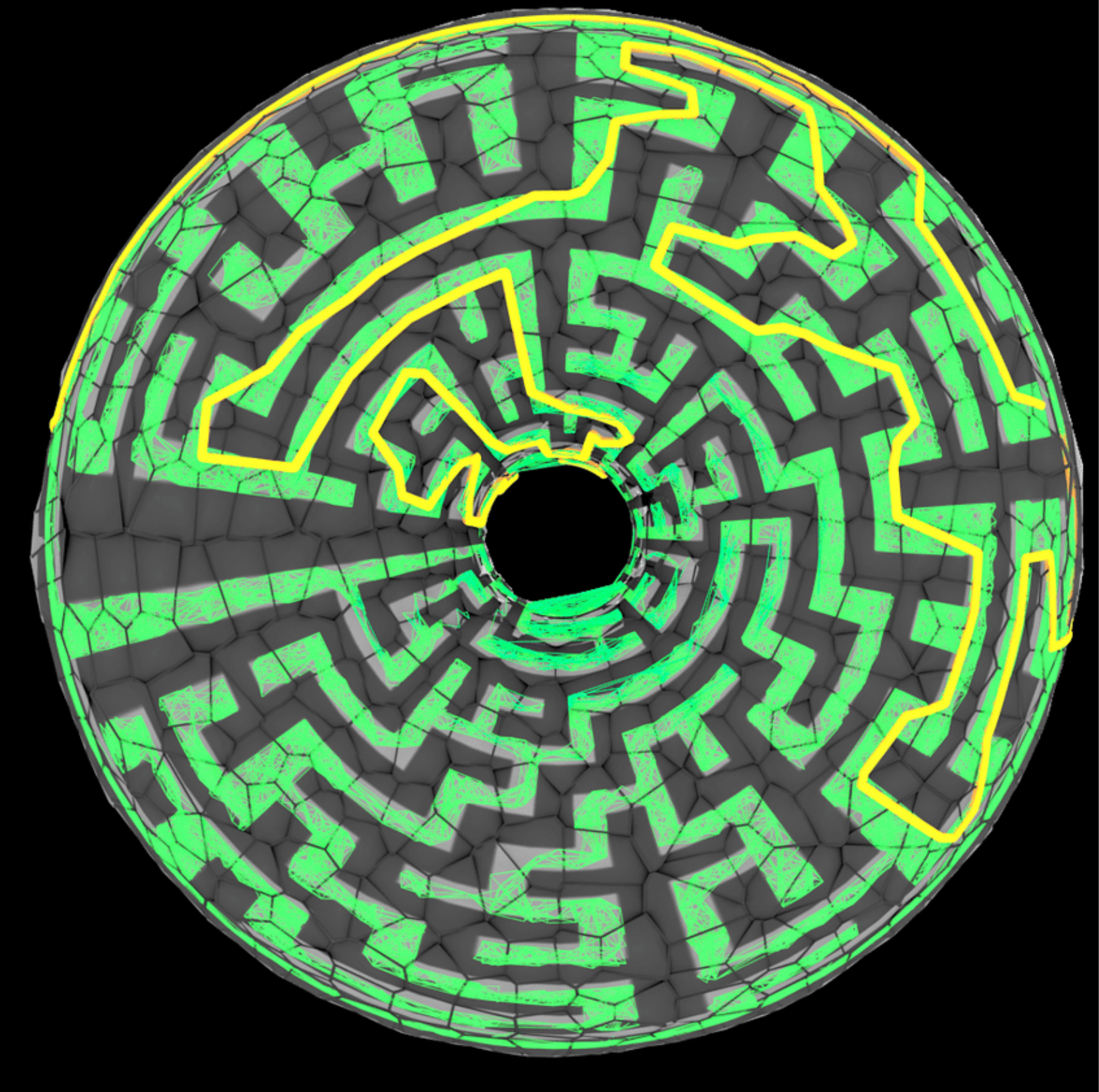}
\end{subfigure}
\begin{subfigure}[t]{\width}
\addSubCaption{Dubin's Car}
\centering
\includegraphics[width=\linewidth]{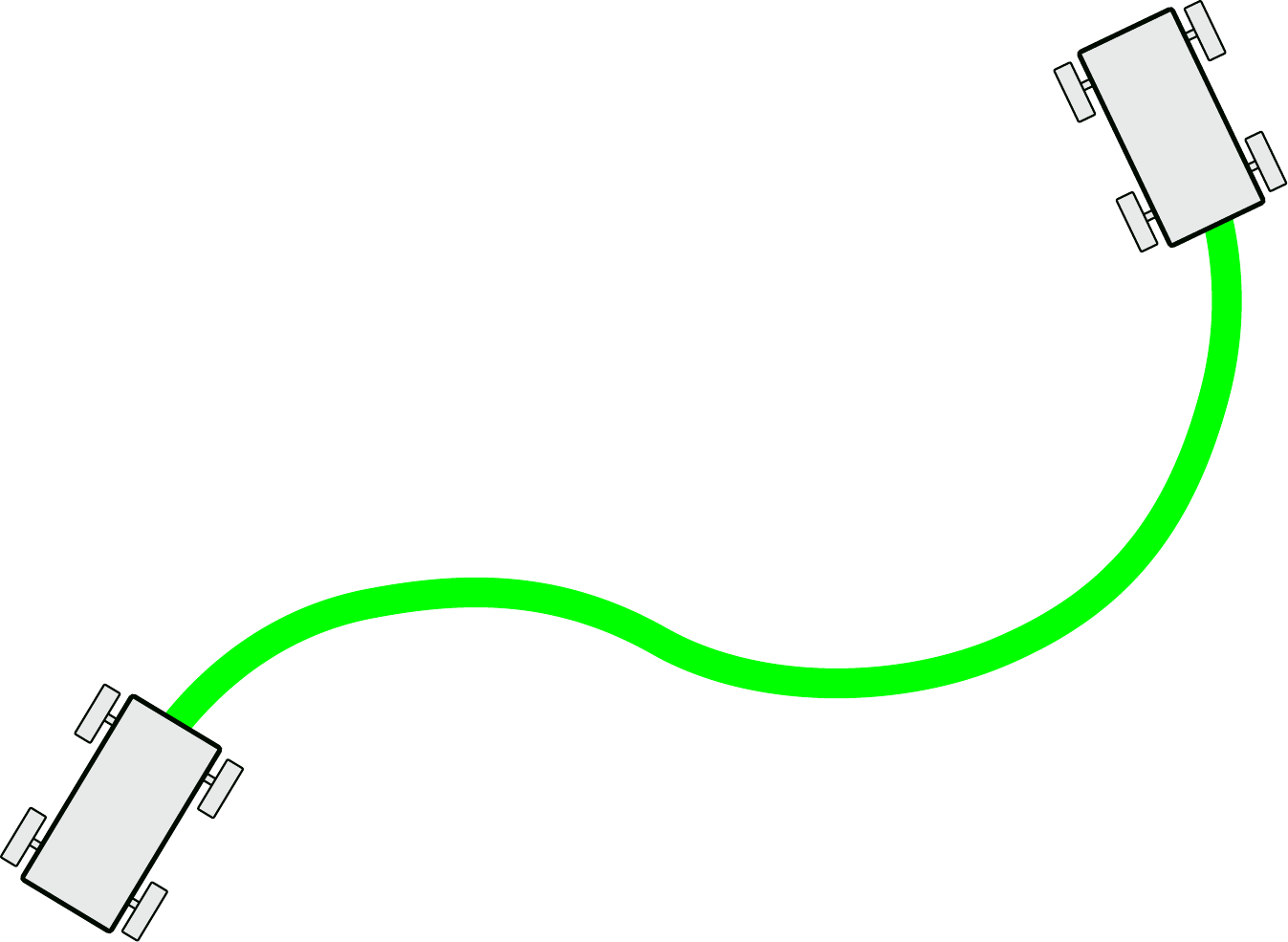}
\end{subfigure}
\begin{subfigure}[t]{\width}
\addSubCaption{Kinematic Car}
\centering
\includegraphics[width=\linewidth]{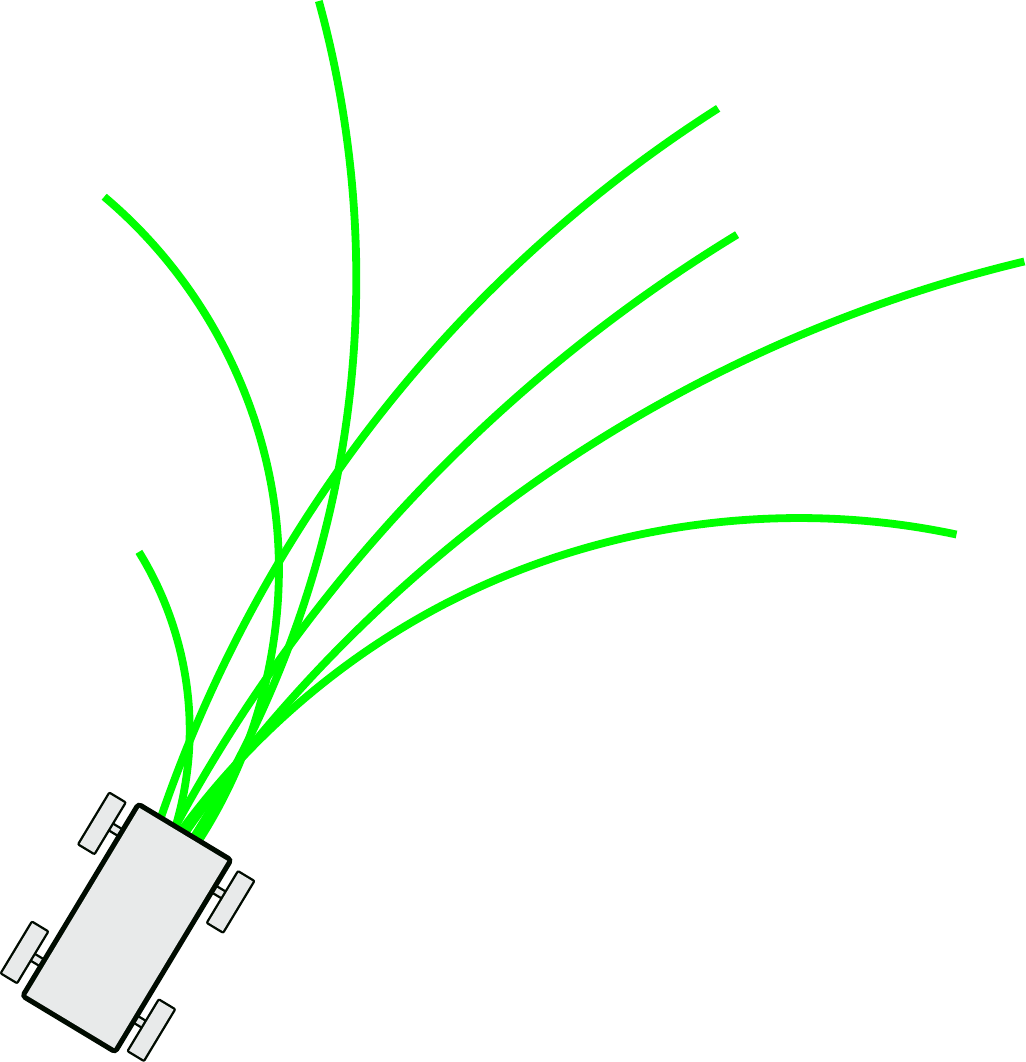}
\end{subfigure}
\begin{subfigure}[t]{\width}
\addSubCaption{UAV}
\centering
\includegraphics[width=\linewidth]{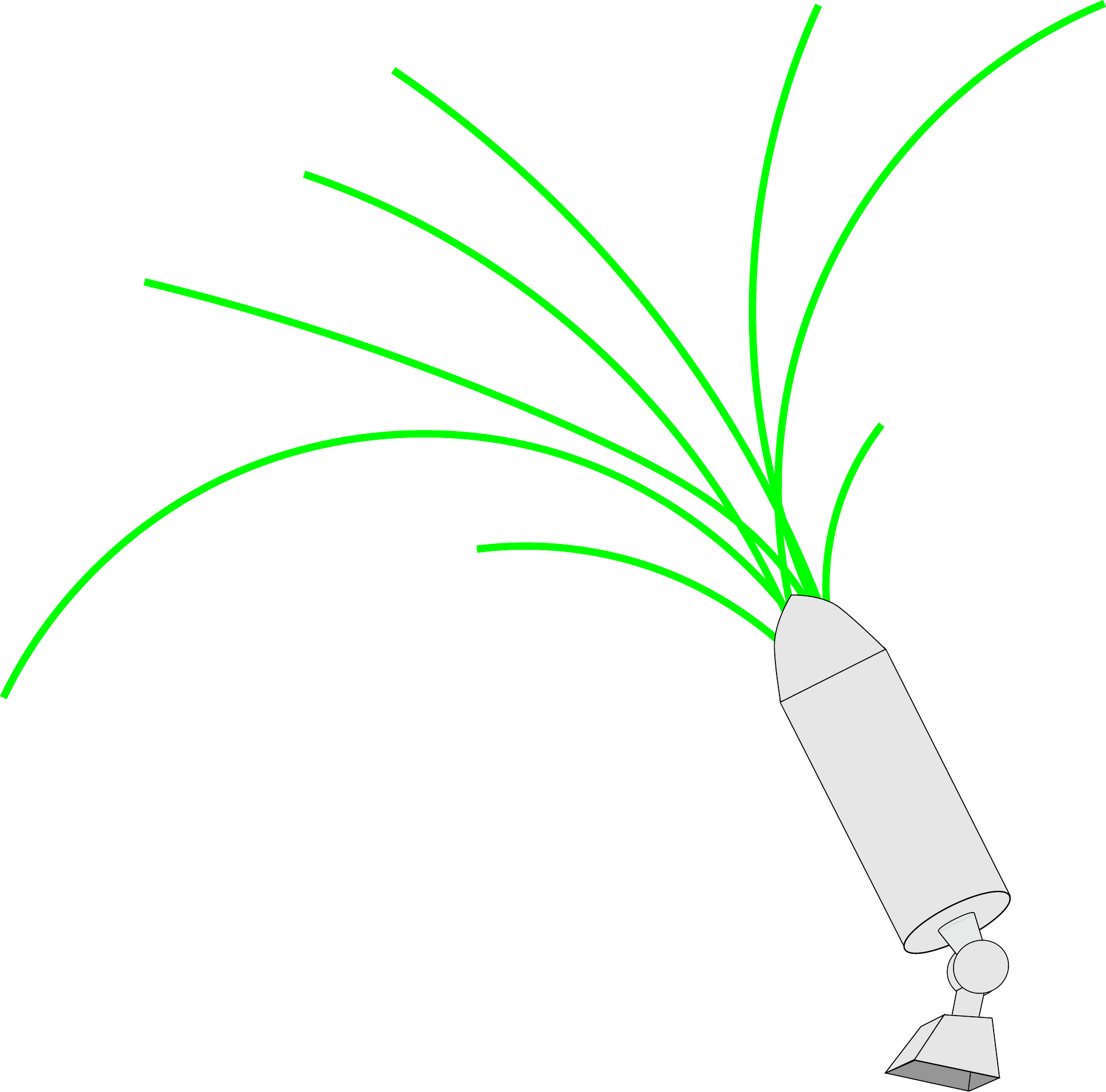}
\end{subfigure}
\def\width{0.32\linewidth}
\begin{subfigure}[t]{\width}
\centering
\includegraphics[width=\linewidth]{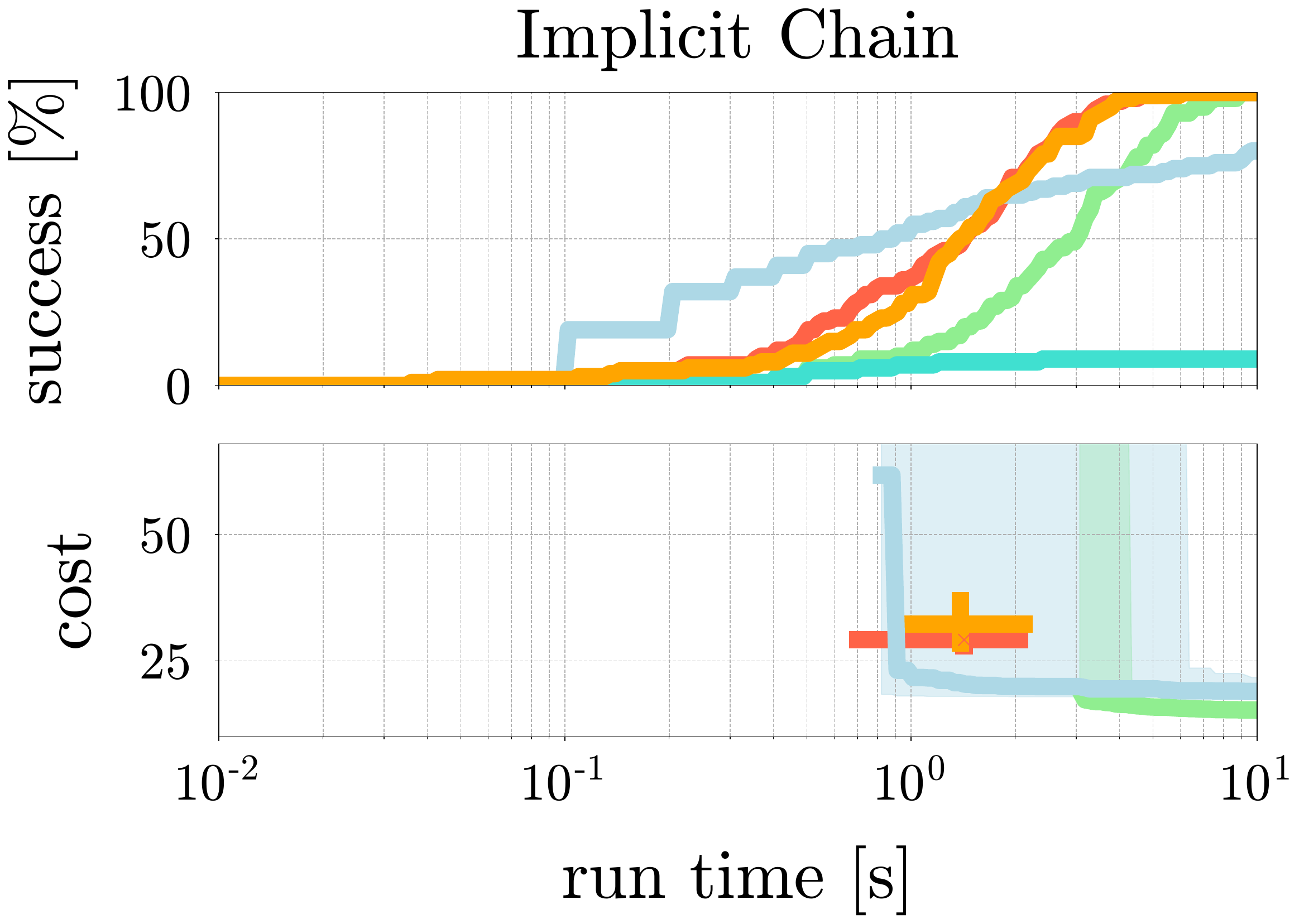}
\end{subfigure}
\begin{subfigure}[t]{\width}
\centering
\includegraphics[width=\linewidth]{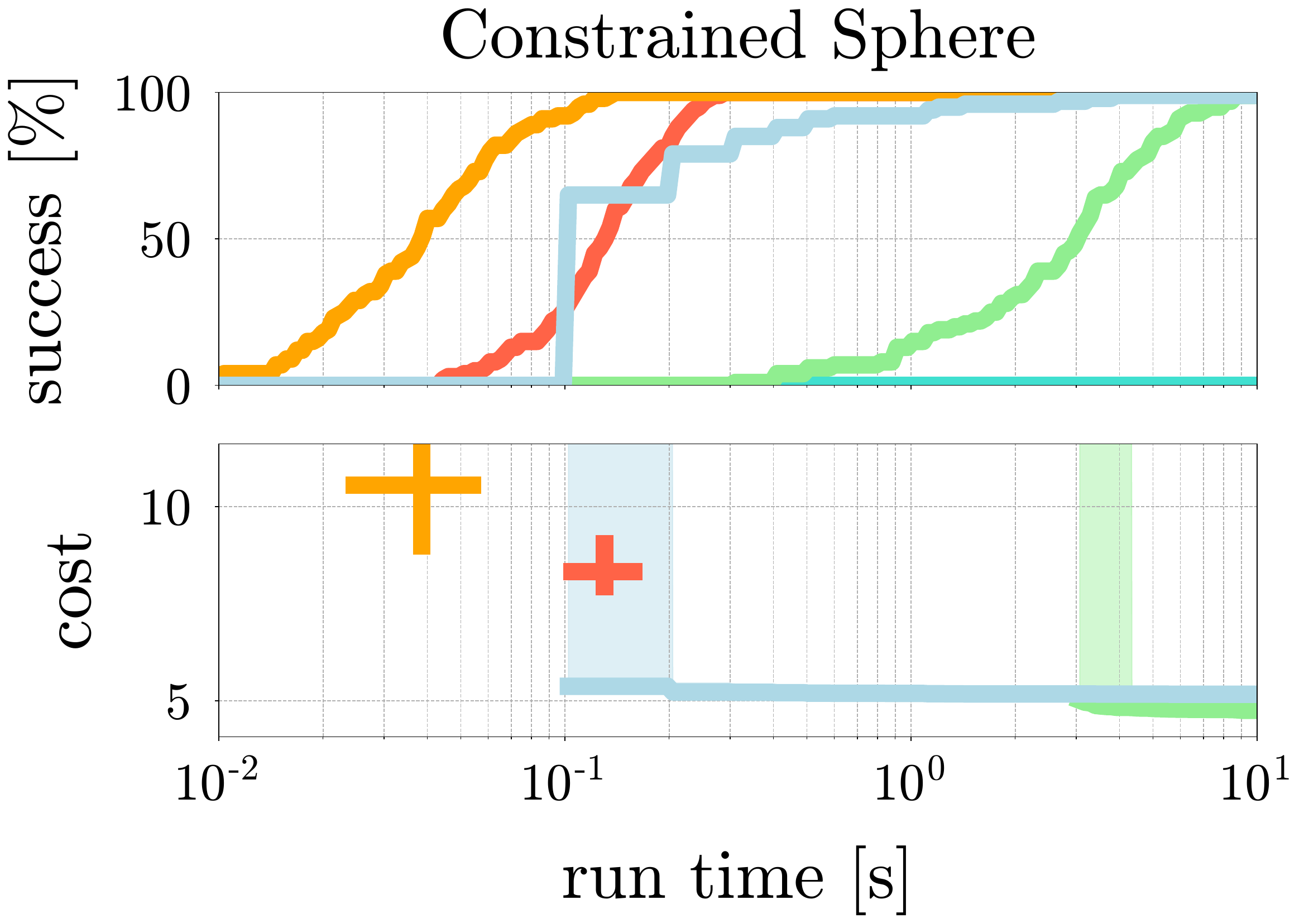}
\end{subfigure}
\begin{subfigure}[t]{\width}
\centering
\includegraphics[width=\linewidth]{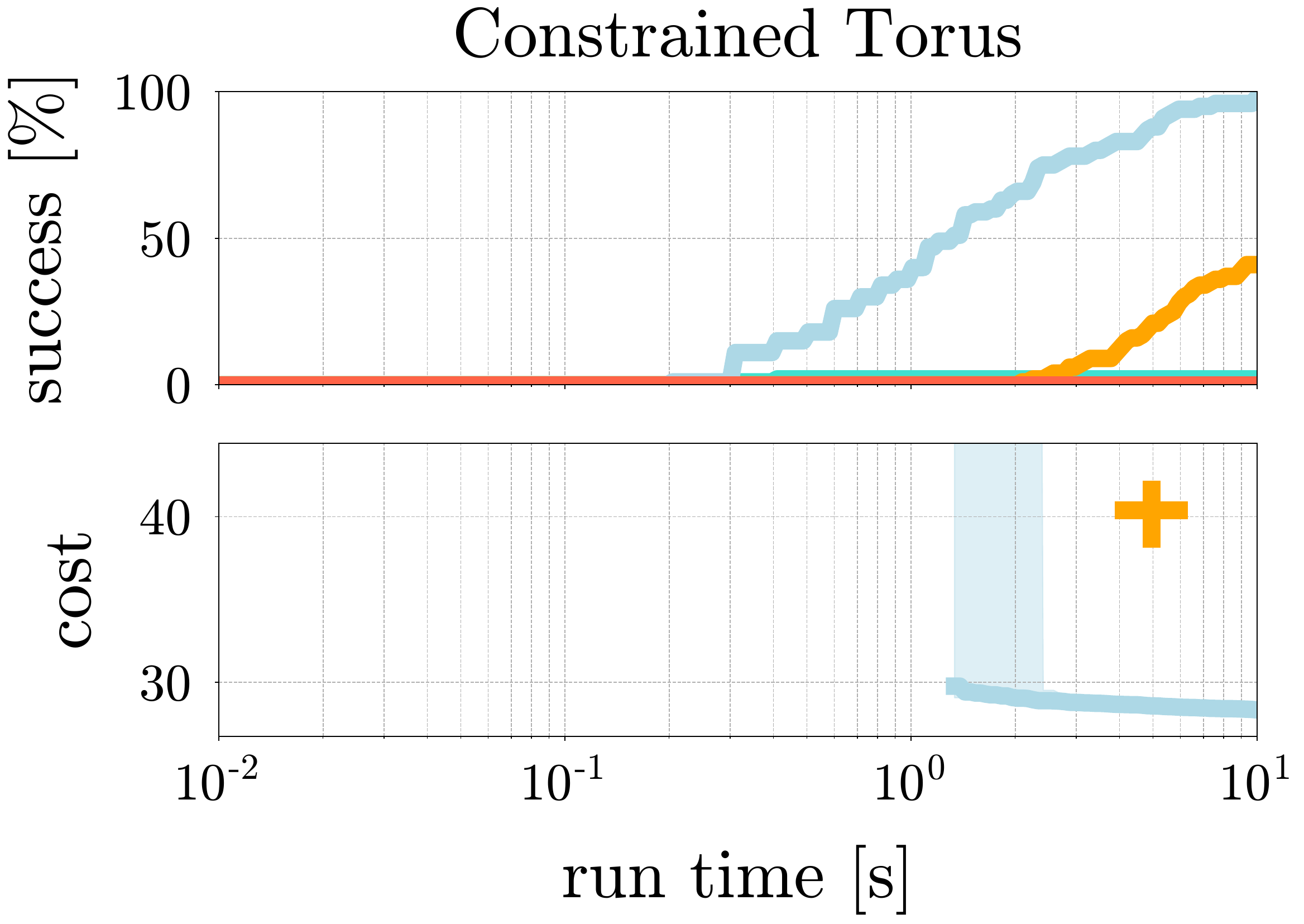}
\end{subfigure}
\begin{subfigure}[t]{\width}
\centering
\includegraphics[width=\linewidth]{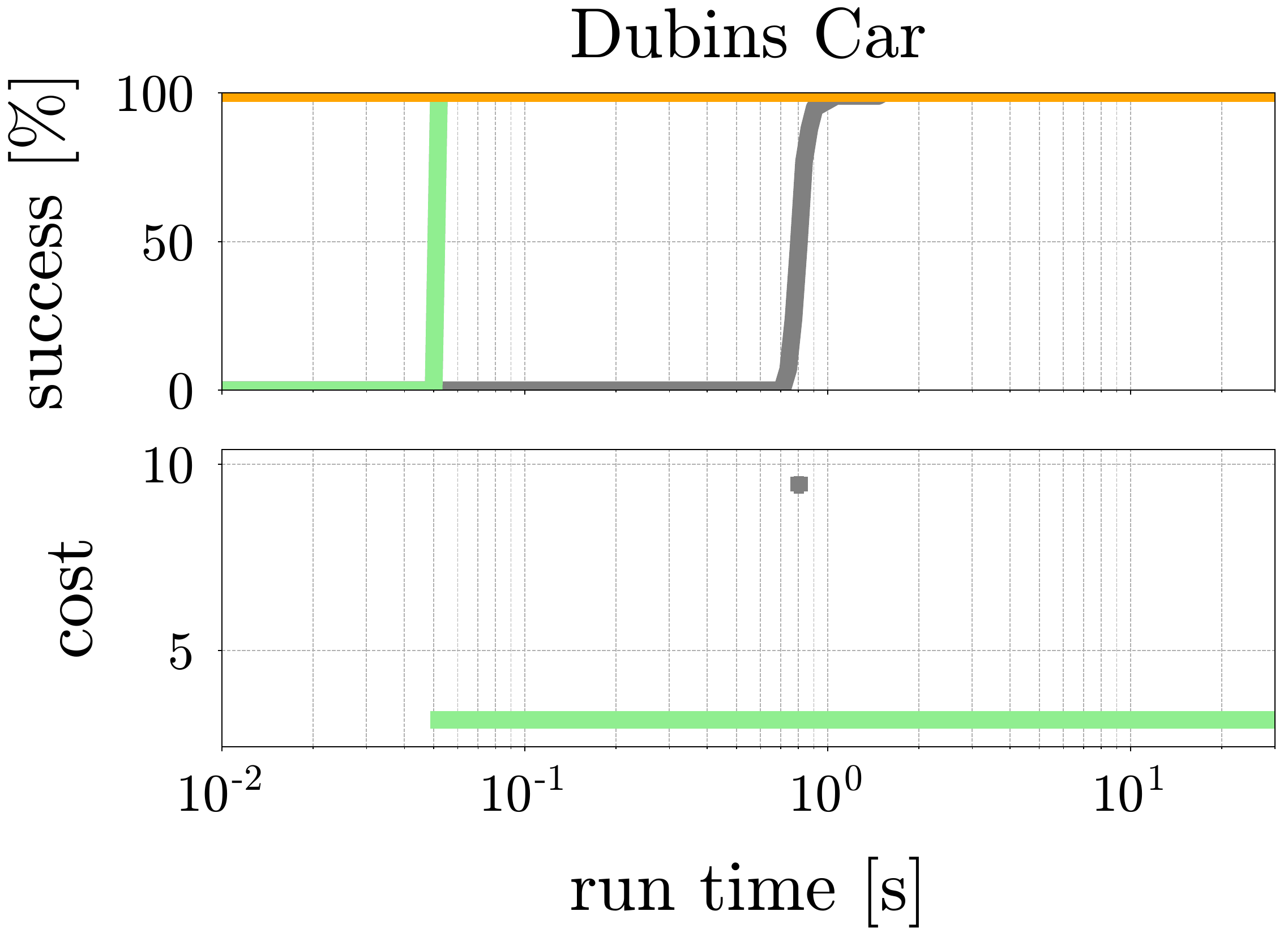}
\end{subfigure}
\begin{subfigure}[t]{\width}
\centering
\includegraphics[width=\linewidth]{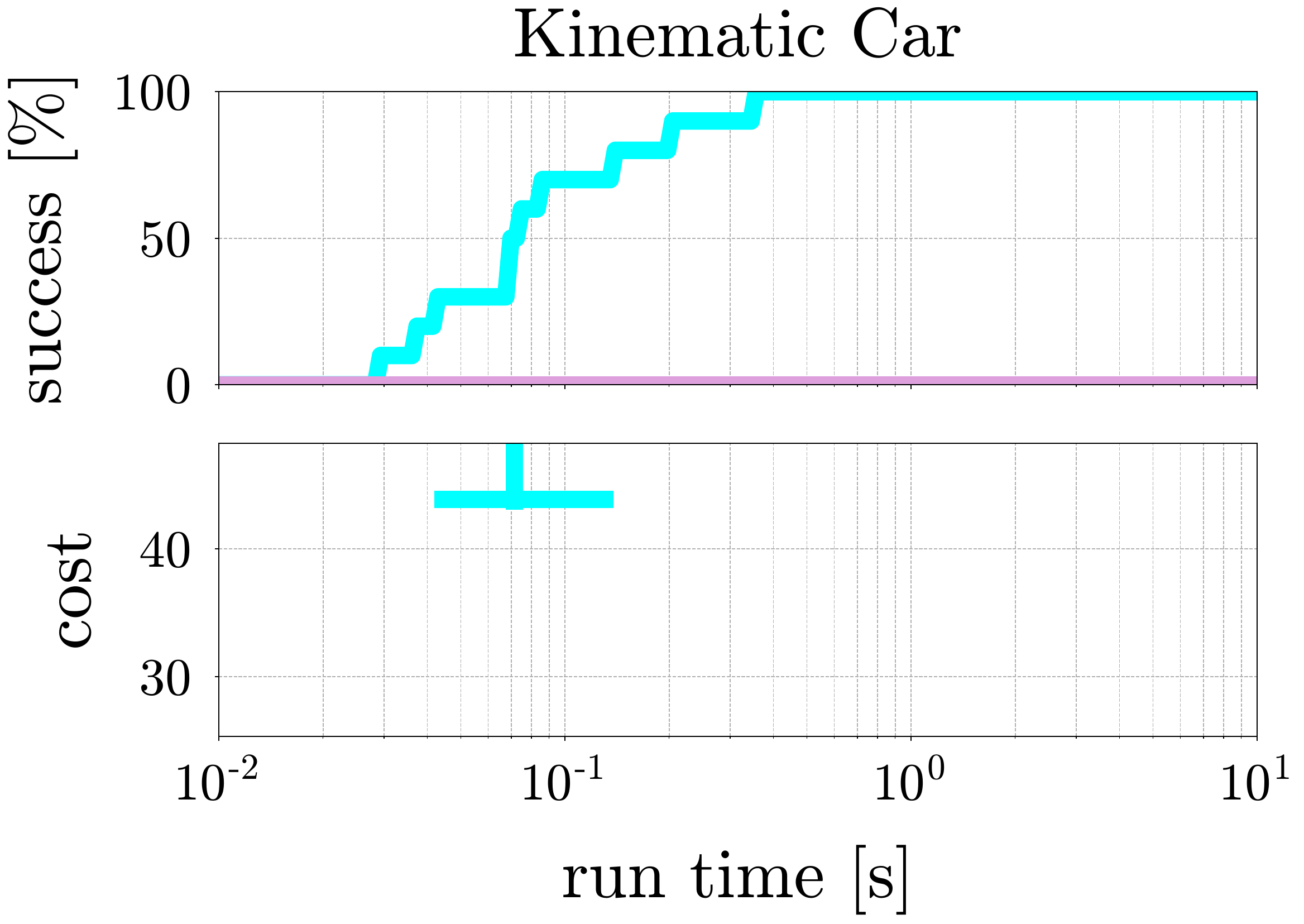}
\end{subfigure}
\begin{subfigure}[t]{\width}
\centering
\includegraphics[width=\linewidth]{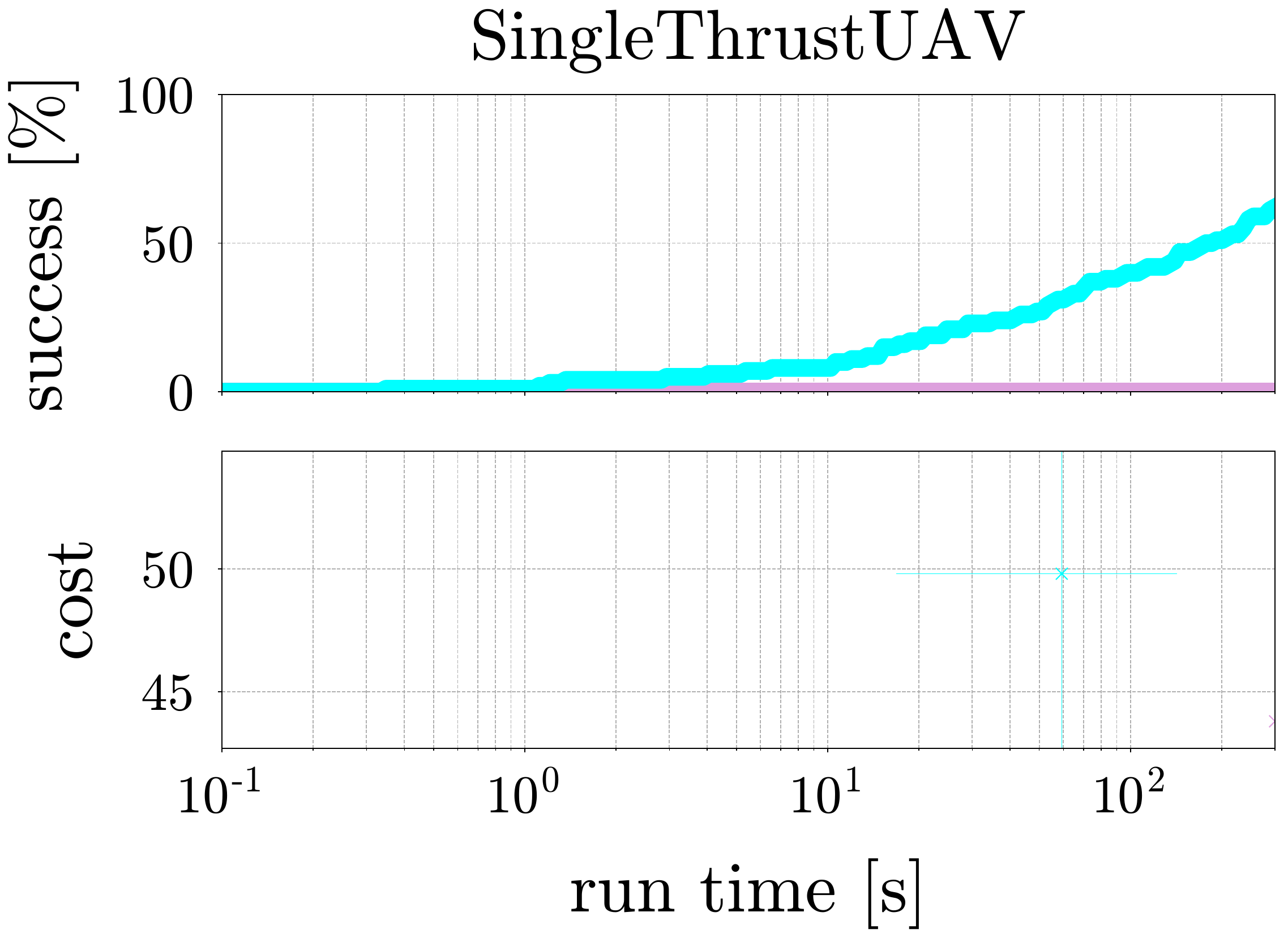}
\end{subfigure}
\caption{Extension Experiments. For the geometric experiments, the planners are \sqbox{crrtconnect}~RRT-Connect, \sqbox{cprm}~PRM, \sqbox{cest}~EST, \sqbox{cbitstar}~BIT*, \sqbox{crrtstar}~RRT*, and \sqbox{cfmt}~FMT. For the dynamic system experiments, the planners are \sqbox{csst}~SST*, \sqbox{ckrrt}~Kinodynamic RRT\label{fig:experiments:extensions}}
\vspace{-1em}
\end{figure*}

%% file: src/08_conclusion.tex
\section{Discussion}

This comparative review provided a reference manual for using sampling-based motion planning algorithms. 
A large-scale evaluation of different planners showed which ones perform best on $24$ challenging scenarios. 
The results indicate that planning algorithms can successfully solve a broad class of problems, even with narrow passages, constraints, or dynamics. However, there is not a single planner that performs best across all problems.
Due to space limitations, a topic not covered in these experiments was the choice of hyper-parameters of sampling-based planners. Hyper-parameters, however, can drastically affect the performance of a motion planning problem. 
This is an active area of research and we refer the reader to \cite{moll2021hyperplan} for a starting point that discusses hyper-parameter tuning for sampling-based planners.
Apart from the comparative evaluations, this review provided a comprehensive overview about state space structures, categories of sampling-based planning algorithms, motion planning extensions, and a comparison to alternative motion generation frameworks. 
This should give researchers and practitioners the tools to make better decisions about which sampling-based planners to use in a specific motion planning scenario.

%% file: src/09_relatedwork.tex
\section{Other Reviews of Interest}

In the last decades, researchers in sampling-based motion planning have published several comprehensive review papers. 
An early account of the history of the field is found in the treatise by Latombe~\cite{Latombe1999}. 
Focusing more on sampling-based approaches, the work by Tsianos et al.~\cite{Tsianos2007} discusses developments of the field during the early 2000's. 
A more comprehensive resource is the work by Elbanhawi and Simic~\cite{Elbanhawi2014} which provides an overview of different planners, together with a set of general primitives for all planners. 

There is also an ever-growing list of review papers focusing on a
specific variant of motion planning or on a specific application field. 
A comprehensive overview of asymptotically-optimal planners is a recent survey by Gammell and Strub~\cite{Gammell2020Survey}. Similarly, but with a focus on heuristic approaches, is the work by Mac et al.~\cite{Mac2016}. Excellent overviews exist also on motion planning inspired methods for molecular simulations~\cite{Gipson2012, AlBluwi2012}, where different required extensions are discussed. For the field of Unmanned Aerial Vehicles (UAVs), the work by Goerzen et al.~\cite{Goerzen2010} provides an overview and discusses more advanced planning problems like surveillance and reconnaissance.
Learning and sampling-based planning is discussed in~\cite{McMahon2022learningreview}.
Excellent reviews for the role of motion planning in task planning can be found in Garrett et al.~\cite{garrett2021integrated}, and Kingston et al.~\cite{kingston2018sampling}, respectively.

While all these works provide good overviews about sampling-based planning and specific areas, this review provides a broader guide for sampling-based motion planning. 
This review is also more application-oriented, in that it also shows the relative performance of popular planners on different problem areas.



